\documentclass{article}

\usepackage[main, final]{neurips_2026}

\usepackage{silence}
\usepackage[utf8]{inputenc} 
\usepackage[T1]{fontenc}    
\usepackage{hyperref}       
\usepackage{url}            
\usepackage{booktabs}       
\usepackage{amsfonts}       
\usepackage{nicefrac}       
\usepackage{microtype}      
\usepackage{xcolor}         
\usepackage{graphicx}
\usepackage{subcaption}
\usepackage{amsmath}
\usepackage{algorithm}
\usepackage[noend]{algpseudocode}
\usepackage{comment}
\usepackage{booktabs}
\usepackage{wrapfig}
\algrenewcommand\algorithmicrequire{\textbf{Input:}}
\algrenewcommand\algorithmicensure{\textbf{Output:}}

\title{Hierarchical Regime-Conditioned Dynamics for Spatiotemporal Graphs}

\author{
  Jeremy Baffou\\
  EPFL\\
  \texttt{jeremy.baffou@epfl.ch} \\
  \And
  Pascal Frossard\\
  EPFL \\
  \texttt{pascal.frossard@epfl.ch} \\
  \And
  Dorina Thanou\\
  EPFL \\
  \texttt{dorina.thanou@epfl.ch} \\
}

\begin{document}

\maketitle

\begin{abstract}
Forecasting the behavior of spatio-temporal systems often requires more than accurate predictions: models should uncover the dynamical modes governing future evolution. Existing temporal graph and latent dynamical models achieve strong forecasting performance, but their latent representations often entangle multiple dynamical patterns, limiting interpretability. Conversely, regime-inference methods explicitly model such modes but typically do not support forecasting over graph-structured systems. We introduce HERON, a self-supervised regime-aware generative framework that unifies graph-structured trajectory generation with regime identification. HERON factorizes the latent state into discrete system-level regimes, capturing global dynamical modes, and continuous entity-level dynamics, yielding a modular regime interface compatible with existing spatio-temporal frameworks. Crucially, the inferred regime actively conditions system dynamics, enabling regime tracking and controlled simulations under alternative dynamical modes. Without regime annotations, HERON jointly learns forecasts and regime dynamics in a self-supervised manner, recovering meaningful modes across controlled synthetic and real-world datasets while maintaining competitive forecasting performance. These results open a path toward interpretable, controllable generative modeling of spatio-temporal dynamics.
\end{abstract}

\section{Introduction}\label{sec:introduction}

Accurate forecasting of the evolution of dynamical systems is essential in a wide range of applications, including healthcare \citep{health1,health2,health3}, finance \citep{finance2,finance1,finance3}, and autonomous motion systems \citep{automotion1,automotion2}. However, when similar observations arise from different underlying mechanisms, predictive accuracy alone is insufficient to characterize the system's dynamics. For example, two patients may display similar tumor growth patterns while belonging to different disease phases, requiring distinct monitoring strategies and interventions \citep{hubert1,hubert2}. In such cases, identifying the underlying dynamical mode provides the context necessary to interpret, monitor, and adapt the forecast. There is therefore a need for models that not only forecast future outcomes, but also track the dynamical patterns that generate them. Such patterns can often be represented as discrete dynamical regimes such as disease phases \citep{stage1,stage2} or market conditions \citep{market1,market2}, providing an interpretable abstraction of otherwise complex interaction-driven processes. However, existing approaches typically address regime inference \citep{regimeinference1,regimeinference2} and trajectory generation \citep{lam-slide,trajectory2,trajectory3} in isolation, leading to a lack of unified frameworks that can simultaneously identify regimes and generate regime-aware dynamics.

In many spatiotemporal settings like multi-agent motion or biological networks, dynamics arise from interactions between entities rather than from independent trajectories, motivating models that explicitly couple temporal evolution with relational structure. Despite recent progress in modeling relational dynamics, existing approaches remain primarily focused on prediction rather than explicit regime characterization. Temporal graph neural networks \citep{tgn1,tgn2,tgn3,ref_roland,ref_evolve_gcn,gnnrecurrent1} and graph state-space models \citep{graphssm,gssm1} address this by combining relational aggregation with temporal memory, achieving strong results in graph-structured forecasting. However, these models are typically optimized for accurate rollouts, with latent states serving as predictive features rather than explicit descriptions of the underlying dynamical mode. This limits their applicability when the question is not only \emph{where will the system go?}, but also \emph{which dynamics govern its evolution?} Conversely, switching dynamical systems and regime-inference methods explicitly model discrete modes \citep{regimeinference1,regimeinference2,changepoint1,changepoint2,changepoint3}, but are rarely designed to generate future trajectories over interacting entities. This creates a gap between forecasting models that encode regimes implicitly in continuous representations and regime-based models that lack graph-structured generative capabilities. This gap raises the central question of this work: can discrete regimes be learned as operational conditioning variables that track the current dynamical mode, govern future graph rollouts, and preserve the forecasting performance of latent trajectory models?

In this work, we introduce \textbf{HERON}\footnote{Code is available at \url{https://github.com/JerBaf/HERON}.}, a self-supervised, regime-aware generative framework for spatio-temporal dynamics that models discrete, interpretable dynamical regimes as latent states governing interacting entities. Concretely, HERON treats each time step as a graph snapshot: given a context of past observations, it forecasts future trajectories over a target topology while inferring, at each step, which of $K$ latent regimes governs the system evolution. HERON employs a hierarchical state representation in which a discrete system-level regime explicitly conditions the evolution of entity-level states. This formulation enables a bidirectional flow of information: local spatio-temporal observations are aggregated to infer a global regime, which in turn governs the generation of future trajectories. By coupling regime inference with autonomous rollout, HERON produces forecasts that are both accurate and dynamically actionable. Crucially, the inferred regime is not merely descriptive but it explicitly parametrizes the transition dynamics via a regime-conditioned mixture-of-experts generator, such that the active regime governs future trajectory generation. Across synthetic and real-world datasets with fixed and time-varying regimes, we show that HERON jointly discovers latent regimes and forecasts trajectories, achieving competitive predictive performance while recovering meaningful and interpretable structure. Furthermore, we show that by intervening on the inferred regimes, HERON can expose plausible alternative future trajectories, enabling decision making through the exploration of multiple possible futures rather than a single predicted trajectory. The key contributions of this work are:
\begin{itemize}
    \item \textbf{Regime-aware framework for spatio-temporal dynamics.}
    We introduce a modular hierarchical latent-state formulation that separates system-level regimes from entity-level dynamics and integrates with existing spatio-temporal backbones.

    \item \textbf{Self-supervised regime-aware forecasting.}
    We show that jointly learning regimes and forecasts through our self-supervised objective recovers meaningful dynamical modes without annotations while maintaining competitive predictive performance.

    \item \textbf{Regime-conditioned rollouts.}
    We use regimes as active conditioning variables for generation, enabling regime tracking and controlled simulations of alternative futures under different dynamical modes rather than optimizing endpoint forecasts alone.
    
\end{itemize}

\section{Problem formulation} \label{sec:problem_formulation}

\begin{figure}[t]
\centering
\includegraphics[width=0.8\textwidth]{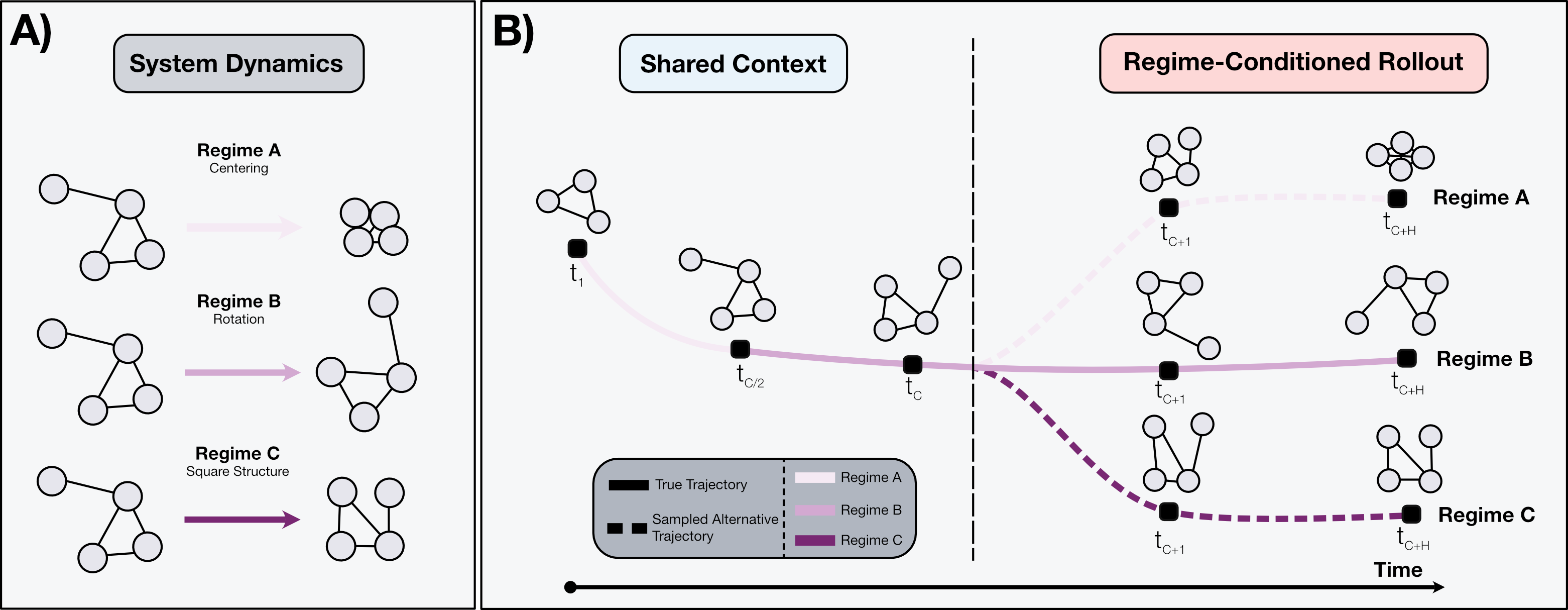 }
\caption{\textbf{Example of regime-conditioned graph rollouts.} \textbf{A)} The system can evolve according to distinct dynamics referred to as regimes. \textbf{B)} Given an observed graph sequence acting as \textit{context}, we forecast future node features while inferring the latent regime trajectory that conditions the \textit{rollout}. The regime exposes which dynamical mode drives the predictions and allows alternative regime-aware futures beyond forecasting alone.}
\label{fig:setup}
\end{figure}

\paragraph{Setup.} We consider a spatio-temporal system of entities 
$\{n_i\} \subset \mathcal{N}$ observed at discrete time steps 
$t \in \{1, \dots, T\}$ with sampling interval $\Delta t$. To capture pairwise relations among entities, we model the system as a sequence of graph snapshots \(\mathcal{G}_t = (\mathcal{V}_t, \mathcal{E}_t, \mathbf{X}_t)\), where \(\mathcal{V}_t\) is a set of \(N_t\) entities, \(\mathcal{E}_t \subseteq \mathcal{V}_t \times \mathcal{V}_t\) is a set of \(M_t\) relations, and \(\mathbf{X}_t = (\mathbf{X}^V_t, \mathbf{X}^E_t)\) collects node and edge features, with \(\mathbf{X}^V_t \in \mathbb{R}^{N_t \times d_V}\) and \(\mathbf{X}^E_t \in \mathbb{R}^{M_t \times d_E}\). The topology \((\mathcal{V}_t, \mathcal{E}_t)\) is given and may vary with \(t\). Node features $\mathbf{X}^V_t$ typically include spatial coordinates and other entity-level covariates.

We model the evolution of the system through a collection of \(K\) \emph{regimes}, each encoding a distinct mechanism by which node and edge features evolve on a given topology. Formally, each regime is represented by a function \(f_k\) that maps the graph history up to time \(t\) to the next feature state: \[f_k:(\mathcal{V}_{1:t}, \mathcal{E}_{1:t}, \mathbf{X}_{1:t}) \mapsto \mathbf{X}_{t+1},\] yielding the set \(\{f_k\}_{k=1}^{K}\). We assume that the number of regimes \(K\) is known and that exactly one regime is active at each time step. When the number of regimes is not known a priori, we select \(K\) from the elbow of held-out predictive NLL. We detail sensitivity to \(K\) and extensions to overlapping regimes in Appendices~\ref{app:k_sensitivity}, \ref{app:k}, and \ref{app:multik}.

\paragraph{Objective.}
Given a \textit{context} \((\mathcal{V}_{1:C}, \mathcal{E}_{1:C}, \mathbf{X}_{1:C})\) of length \(C\) and a target topology \((\mathcal{V}_\star, \mathcal{E}_\star)\), we aim to (i) learn a set of regimes \(\{f_k\}_{k=1}^{K}\), and (ii) jointly infer the regime trajectory, i.e., the sequence of active regimes \((r_t)_{t=1}^{C+H}\) with \(r_t \in \{1,\dots,K\}\), while generating the future features \(\mathbf{X}_{C+1:C+H}\) over a \textit{horizon} of length \(H\) on the target topology. By default, \((\mathcal{V}_\star, \mathcal{E}_\star) = (\mathcal{V}_C, \mathcal{E}_C)\), i.e.\ the last observed topology. Figure~\ref{fig:setup} illustrates this setting, showing how future graph features are rolled out while jointly inferring the latent regime trajectory that conditions their evolution.

\section{Unifying regime inference and forecasting on spatio-temporal graphs} \label{sec:method}

Building on the spatio-temporal setting introduced in Section~\ref{sec:problem_formulation}, we now present \textbf{HERON}, (HiErarchical Regime-cOnditioned dyNamics), a hierarchical generative graph state-space framework for joint regime inference and trajectory forecasting. HERON combines a discrete system-level latent variable capturing global regimes with continuous entity-level states for local dynamics. At each time step, the model infers the current regime from graph-temporal context, uses it to condition the dynamics, and predicts future system features. We detail the state representation, architecture, and training procedure below.

\subsection{State representation} \label{sec:state_representation}

To capture both system-level regimes and entity-specific dynamics, we model a hierarchical latent state separating global and node-level processes. At each level, the state combines a deterministic recurrent component, which accumulates temporal evidence, with a stochastic component that captures uncertainty and promotes diverse trajectories. Formally, at time step $t$, the hierarchical state decomposes as follows:

\textbf{System level}. A deterministic recurrent state $h_t^G \in \mathbb{R}^{d_H}$ captures global temporal context, while a discrete stochastic variable $r_t$, parameterized by a probability distribution $\pi_t$ over $K$ regimes with realization represented as one-hot vector, encodes the active mode of collective behavior.

\textbf{Entity level}. Each node $v_i$ maintains a deterministic recurrent state $h_t^i \in \mathbb{R}^{d_h}$ and a continuous stochastic state $z_t^i \in \mathbb{R}^{d_z}$, jointly encoding entity-specific dynamics.

The full hierarchical state at time $t$ is thus:

\[h_t = \bigl(h_t^G,\; \{h_t^i\}_{i \in \mathcal{V}_t}\bigr), \qquad s_t = \bigl(r_t,\; \{z_t^i\}_{i \in \mathcal{V}_t}\bigr).\]

While recurrent state-space models (RSSMs) \citep{planet,dreamerv2,dreamerv3} also combine deterministic and stochastic components, they use a single latent state. This limits their ability to model structured dynamics.  In contrast, our hierarchy explicitly disentangles global regimes from entity-level dynamics, enabling structured modeling  that cannot be captured by a single latent state.

\subsection{Hierarchical regime-conditioned architecture} \label{sec:architecture}

\begin{figure}[ht]
\centering
\includegraphics[width=0.8\textwidth]{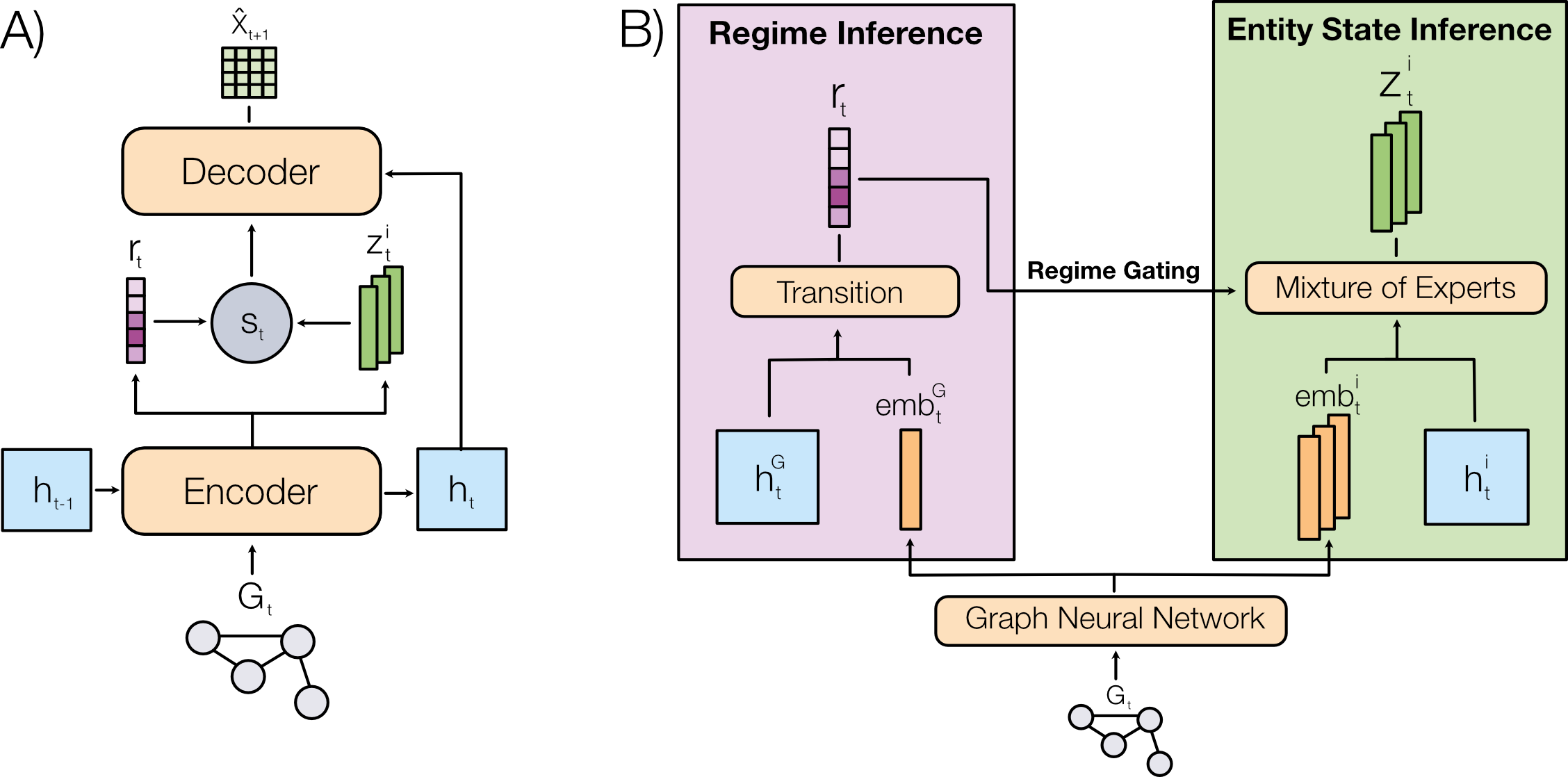}
\caption{\textbf{Model architecture.} \textbf{A) Framework:} At each time step, the model first updates the deterministic recurrent state $h_t$. Conditioned on this state, the encoder infers the regime $r_t$ and the entity latents $z_t^i$. Together with $h_t$, they are then used by the decoder to predict the next-step features $X_{t+1}$. \textbf{B) Encoder:} The encoder proceeds hierarchically. A GNN first extracts node- and graph-level embeddings (${\mathrm{emb}}_t^i$, ${\mathrm{emb}}_t^G$), which are used to infer the regime ${r}_t$ via a transition module. This regime then conditions a mixture-of-experts (MoE) to produce the entity-level latent states ${z}_t^i$.}

\label{fig:architecture}
\end{figure}

HERON follows a hierarchical encoder-decoder architecture as shown in Figure \ref{fig:architecture}. At each time step, recurrent state updates summarize past dynamics at both the system and entity levels. These states are then combined with graph information to produce node- and graph-level embeddings. From these embeddings, the encoder infers a system-level regime together with regime-conditioned latent states at the entity level, which are in turn passed to the decoder to predict next-step node and edge features. We detail these components below following the progression from recurrent state updates to relational embeddings, hierarchical regime inference, and decoding. Figure \ref{fig:architecture} summarizes the complete HERON forward-pass.

\subsubsection{Inference paths} \label{sec:inference-paths}

HERON represents the stochastic state $s_t=\bigl(r_t,\; \{z_t^i\}_{i \in \mathcal{V}_t}\bigr)$ via two distributions over the same variables. The \textbf{posterior} $q_\phi(s_t \mid h_t, X_t, E_t)$ conditions on the current node and edge features and topology and is used over the context window. The \textbf{prior} $p_\psi(\hat{s}_t \mid h_t, E_t)$ conditions only on the deterministic state and topology and is used during rollout, when features are unavailable. We use hats throughout to denote prior quantities. The two paths share the recurrent state updates and the decoder; only the relational embeddings and the stochastic transition heads are duplicated as separate $\phi$- and $\psi$-modules. This dual-path design enables latent rollouts to be generated directly from test-time inputs, rather than feeding predicted features back into a single encoder, which can lead to error accumulation. The training objective (Section \ref{sec:training_procedure}) couples the two distributions through a Kullback–Leibler (KL) term, so that the prior remains consistent with observation-conditioned inference. We describe below the encoder from the posterior path; the prior follows the same structure with its own modules and without $X_t$ as input. See the Appendix \ref{app:posterior_prior} for more details.

\subsubsection{Encoder}

The encoder updates the deterministic states and infers the stochastic variables of the current hierarchical state $(h_t, s_t)$, as defined in Section~\ref{sec:state_representation}, from two complementary sources of information: temporal memory carried by the recurrent states, and relational evidence extracted from the current graph snapshot when available. Concretely, the deterministic states $h_t$ summarize past dynamics, while the embeddings $\mathrm{emb}_t=(\mathrm{emb}_t^i, \mathrm{emb}_t^G)$ provide relational summaries of the current graph. The encoder therefore first propagates the deterministic recurrent states forward in time, and then infers the stochastic latent variables through a hierarchical pipeline.

\textbf{Recurrent state update}. We first propagate temporal information forward through the deterministic part of the state. At both levels of the hierarchy, gated recurrent units (GRUs) \citep{GRU} update the recurrent states from the previous latent state:
\begin{equation}
h_t^i = GRU_{\theta}^\mathcal{V}(h_{t-1}^i,z_{t-1}^i), \qquad h_t^G = GRU_{\theta}^G(h_{t-1}^G,r_{t-1}), 
\end{equation}
where $GRU_{\theta}^\mathcal{V}$ and $GRU_{\theta}^G$ denote two distinct GRU transition functions for the entity- and system-level recurrent states, respectively.

\textbf{Relational embeddings}. Conditioned on the recurrent states, a GNN\footnote{In this work, we instantiate the relational encoder with GNNs to model interactions among entities. However, HERON is agnostic to the specific relational architecture, meaning that any encoder that maps a set of interacting entities to relational representations, such as Set Transformers or point-cloud architectures, can be used in place of the GNN.} computes node embeddings ($\mathrm{emb}_t^i$) that summarize the current local relational configuration. Then, an attention-based pooling operator \citep{attpooling} aggregates node embeddings into a graph-level representation ($\mathrm{emb}_t^G$) :

\begin{equation} 
\mathrm{emb}_t^i = \mathrm{GNN}_\phi\!\left(h_t^i,\, \mathbf{X}_t,\, \mathcal{E}_t\right), \qquad \mathrm{emb}_t^G = \mathrm{Pool}_\phi\!\left(\{\mathrm{emb}_t^i\},\, \mathcal{E}_t\right).
\end{equation}

\begin{wrapfigure}[23]{r}{0.4\textwidth}
\centering
\includegraphics[width=\linewidth]{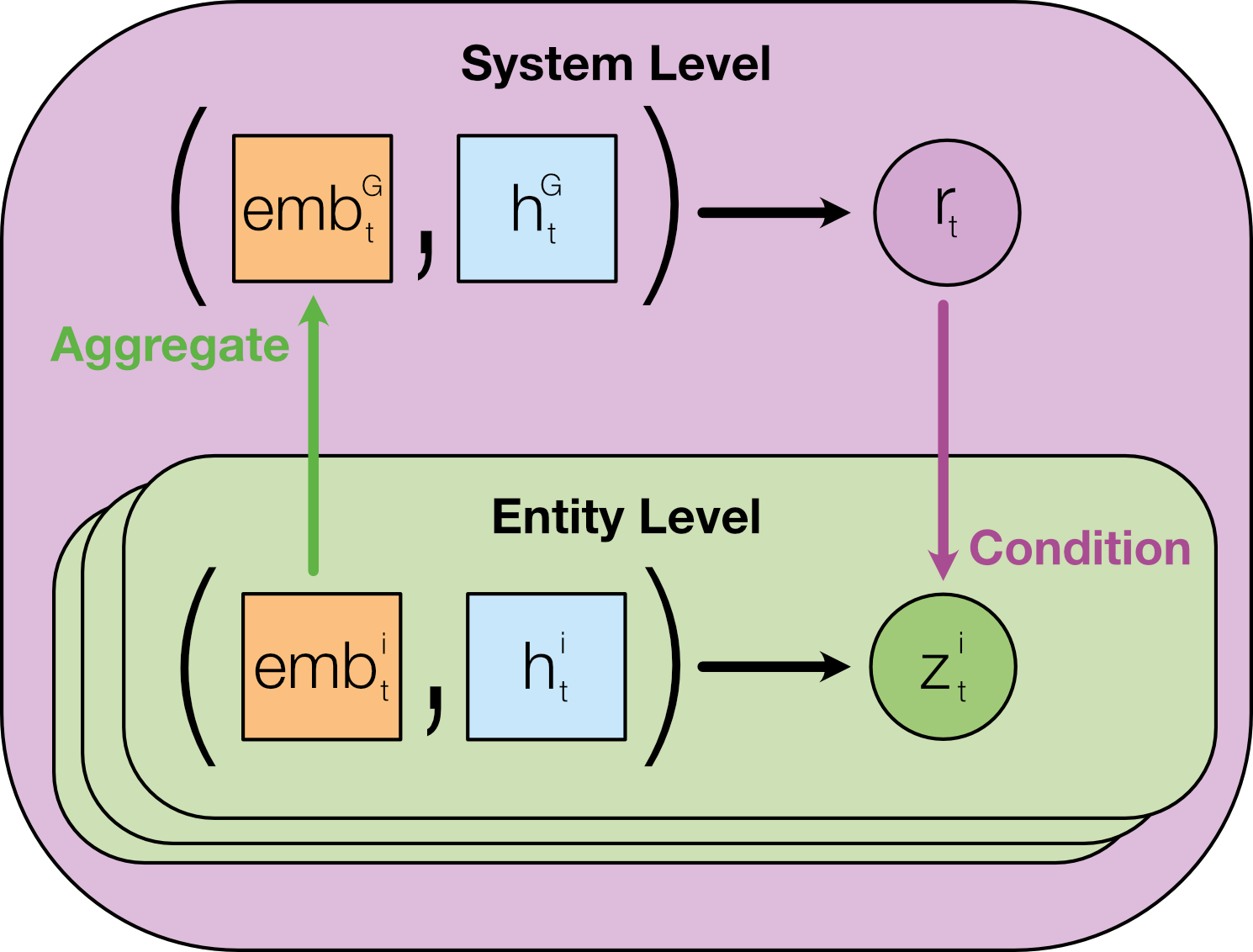}
\caption{\textbf{Hierarchical state representation.} Entity-level embeddings are \textit{aggregated} and combined with the global state $h_t^G$ to infer a latent regime $r_t$ that governs system-wide dynamics. This regime then \textit{conditions} a MoE module to produce entity-level latent states $z_t^i$, linking global behavior and individual evolution.}
\label{fig:state}
\end{wrapfigure}

\textbf{Regime inference}. We model the discrete latent regime variable $r_t$ as a categorical distribution within a variational framework. Sampling is enabled via the Gumbel--Softmax reparameterization \citep{gumbel1,gumbel2} with a straight-through estimator \citep{straight}, which approximates hard regime assignments while preserving a soft representation for gradient-based optimization. The regime transitions are conditioned on the graph-level recurrent state and the corresponding embedding:
\begin{equation}
 r_t\sim q_\phi\!\left(r_t \mid h_t^G,\, \mathrm{emb}_t^G\right).
\end{equation}

\textbf{Regime-conditioned entity state inference}. The inferred regime  explicitly modulates entity-level dynamics rather than acting as a passive system-level label. To enforce this, we use the regime to gate a mixture of experts (MoE) \citep{moe1,moe2} during entity-level stochastic state inference, with one expert assigned to each discrete regime dynamics. Each expert can then be seen as one of the dynamics modes $f_k$ described in Section \ref{sec:problem_formulation}. This induces regime-specific dynamical mappings that generate the node-level latent states, tightly coupling collective behavior with individual dynamics:
\begin{equation}
z_t^i \sim q_\phi\!\left(z_t^i \mid h_t^i,\, \mathrm{emb}_t^i,\, r_t\right).
\end{equation}

\textbf{Prior path.} Starting from the updated deterministic states $h_t$, the prior path embeddings are computed from recurrent states and topology alone:
\begin{equation}
\widehat{\mathrm{emb}}_t^i =
\mathrm{GNN}_{\psi}\!\left(h_t^i,{\mathcal{E}}^\star\right),
\qquad
\widehat{\mathrm{emb}}_t^G =
\mathrm{Pool}_{\psi}\!\left(\{\widehat{\mathrm{emb}}_t^i\}_{i\in{\mathcal{V}}^\star},{\mathcal{E}}^\star\right).
\end{equation}
Then, the hierarchical latent states are sampled sequentially in two steps: the global regime is first sampled from the graph-level prior, after which each entity-level latent state is sampled from its node-level prior conditioned on the sampled regime, i.e.,
\begin{equation}
\begin{aligned}
\hat r_t
&\sim
p_{\psi}\!\left(\hat r_t \mid h_t^G,\widehat{\mathrm{emb}}_t^G\right), \quad
\hat z_t^i
&\sim
p_{\psi}\!\left(\hat z_t^i \mid h_t^i,\widehat{\mathrm{emb}}_t^i,\hat r_t\right),
\qquad i\in{\mathcal{V}}^\star.
\end{aligned}
\end{equation}

Overall, the hierarchical inference pipeline, illustrated in Fig. \ref{fig:state}, aggregates local spatio-temporal signals into a global regime and propagates this information back to entity-level states, enabling explicit interaction between collective and individual dynamics. A summary of the algorithmic procedure is given in Algorithm \ref{alg:HERON}.

\textbf{Integration to existing frameworks}. Importantly, this formulation is agnostic to the underlying temporal graph or relational backbone. It provides a lightweight, regime-aware interface for latent relational models with entity-level temporal states: node-level latent variables are reused as entity states, while a system-level discrete variable, conditions entity-level transitions or decoding. GraphSSM ~\citet{graphssm} provides a natural example of this interface, as it models entity-level latent dynamics through graph-based relational mixing. HERON extends it with a graph-level regime inference module, regime-conditioned entity transitions, and the self-supervised objective of Section~\ref{sec:training_procedure}, equipping the base model with explicit regime tracking and regime-conditioned rollouts while preserving its original inductive biases. We detail the corresponding implementation in Appendix \ref{app:graphssm-HERON}.

\subsubsection{Decoder}

Given the inferred hierarchical state $(h_t, s_t)$, the decoder $g_\theta$ predicts the next set of features $\mathbf{X}_{t+1}$. We use a probabilistic decoder to parameterize the conditional uncertainty over future spatio-temporal dynamics. Concretely, a multi-layer perceptron (MLP) \citep{mlp} maps $(h_t,s_t)$ to the parameters of a Gaussian predictive distribution with diagonal covariance:
\[
\mathbf{X}_{t+1} \sim g_\theta\!\Bigl(\mathbf{X}_{t+1}\mid h_t, s_t\Bigr)
= \mathcal{N}\!\Bigl(\mu_\theta\bigl(h_t,s_t\bigr),\, \mathrm{diag}\!\left(\sigma_\theta^2(h_t,s_t)\right)\Bigr),
\]
 where $\mu_\theta(\cdot)$ denotes the predicted mean and $\sigma_\theta^2(\cdot)$ the predicted marginal variances. The complete encoder--transition--decoder pipeline is summarized in Fig.~\ref{fig:architecture}.

\subsection{Training procedure}\label{sec:training_procedure}

HERON is trained end-to-end in a self-supervised manner, enabling regime discovery directly from spatio-temporal data without annotations. The training objective combines complementary components: accurate features forecasting, alignment between posterior and prior to support autonomous rollout, and structural regularization to prevent degenerate or uninformative regime assignments.

\textbf{Prediction loss}. The loss is defined as a negative Gaussian log-likelihood applied to both the posterior and prior paths, i.e.,  
\[ \mathcal{L}_{\mathrm{pred}} = -\ln g_\theta(\mathbf{X}_{t+1} \mid h_t, s_t) - \gamma \ln g_\theta(\mathbf{X}_{t+1} \mid h_t, \hat{s}_t). \]
 This encourages accurate predictions and consistency between observed and autonomous rollouts.

\textbf{Posterior-Prior alignment}. We align the observation-conditioned posterior with the observation-free prior so that latent trajectories inferred during training can be reproduced at inference time. Following the KL balancing principle  \citep{dreamerv2}, we decompose this KL objective into two asymmetric terms. The dynamics loss $\mathcal{L}_{\mathrm{dyn}}$ trains the prior to match a stopped-gradient posterior ($\mathrm{sg}(q_\phi)$), while the representation loss $\mathcal{L}_{\mathrm{rep}}$ regularizes the posterior toward a stopped-gradient prior ($\mathrm{sg}(p_\psi)$): 
\[\mathcal{L}_{\mathrm{dyn}} = \mathrm{KL}\!\left[\mathrm{sg}\!\left(q_\phi(s_t \mid h_t, \mathbf{X}_t)\right) \,\|\, p_\psi(s_t \mid h_t)\right],\qquad\mathcal{L}_{\mathrm{rep}} = \mathrm{KL}\!\left[q_\phi(s_t \mid h_t, \mathbf{X}_t) \,\|\, \mathrm{sg}\!\left(p_\psi(s_t \mid h_t)\right)\right].\]
Together, these terms train the prior without observation access while preventing the posterior from drifting into regions the prior cannot reproduce.

\textbf{Structural loss.} We regularize the regime pathway with three terms enforcing balanced regime usage, sharp assignments, and decorrelated experts. Let $Q_t^{(b)} = q_\phi(r_t \mid h_{t,b}^G, \mathrm{emb}_{t,b}^G)$ be the posterior regime distribution for sample $b$, and $z_{k,t}^i$ the output of expert $k$ for entity $i$ at time $t$. We define
\[ \mathcal{L}_{\mathrm{struct}} = \mathcal{L}_{\mathrm{bal}} + \mathcal{L}_{\mathrm{ent}} + \mathcal{L}_{\mathrm{spec}} \]
where individual loss components can be writen as the following
\begin{equation}
\mathcal{L}_{\mathrm{bal}} = \mathrm{KL}\!\left[\mathbb{E}_{b}[Q_t^{(b)}] \,\|\, \mathcal{U}(K)\right], \quad
\mathcal{L}_{\mathrm{ent}} = \mathbb{E}_{b}\!\left[\mathcal{H}[Q_t^{(b)}]\right], \quad
\mathcal{L}_{\mathrm{spec}} = \sum_{i \in V_t} \sum_{j \neq k} \frac{\langle z_{j,t}^i, z_{k,t}^i \rangle^2}{\lVert z_{k,t}^i \rVert_2^2 + \epsilon},
\end{equation}
where $\epsilon>0$ ensures numerical stability. $\mathcal{L}_{\mathrm{bal}}$ matches the batch-averaged regime distribution to the uniform prior $\mathcal{U}(K)$, preventing collapse onto a few experts; $\mathcal{L}_{\mathrm{ent}}$ minimizes per-instance regime entropy to encourage confident assignments compatible with straight-through gating; $\mathcal{L}_{\mathrm{spec}}$ minimizes pairwise expert-output projections per entity, encouraging orthogonal expert representations.

\textbf{Final formulation}. The overall objective minimizes the four loss terms over the training window $W=C+H$:
\[ \mathcal{L}(\theta, \phi, \psi) = \mathbb{E}_{q_\phi, p_\psi}\!\left[\frac{1}{W}\sum_{t=1}^{W}\Bigl(\beta_{\mathrm{pred}}\mathcal{L}_{\mathrm{pred}} + \beta_{\mathrm{dyn}}\mathcal{L}_{\mathrm{dyn}} + \beta_{\mathrm{rep}}\mathcal{L}_{\mathrm{rep}} + \beta_{\mathrm{struct}}\mathcal{L}_{\mathrm{struct}}\Bigr)\right].\]

This self-supervised objective enables regime discovery directly from spatio-temporal data while supporting stable rollout and structured modeling of latent dynamics.

\section{Experimental setup} \label{sec:experiments}

\textbf{Baselines} We evaluate HERON against baselines spanning diverse inductive biases for spatio-temporal trajectory prediction. We first include two persistence baselines, namely $\emph{Static}: \hat{x}_{t+1} = x_t$ and $\emph{Velocity}: \hat{x}_{t+1} = x_t + (x_t-x_{t-1})$. We include S4 \citet{s4} as a non-relational state-space baseline. To assess whether performance gains follow from discrete switching alone or MoE conditioning, we compare against rSLDS \citet{rslds}, RED-SDS \citet{redsds} and TESTAM \citet{testam}. To evaluate the role of relational modeling, we further consider LaM-SLidE \citet{lam-slide} and GraphSSM \citet{graphssm}, which extend state-space models with relational inductive or graph-based biases, and EvolveGCN \citet{ref_evolve_gcn} as a temporal GNN baseline. Finally, we evaluate HERON on top of GraphSSM and compare its regime inference against a hidden Markov model (HMM) applied to the GraphSSM embeddings for regime retrieval. Implementation details are provided in Appendix \ref{app:graphssm-HERON}.

\textbf{Evaluation metrics} We evaluate HERON along two axes: \textbf{1) Forecasting:} We measure forecasting quality using Average Displacement Error (ADE) \citet{ade1} and minimum Final Displacement Error (minFDE) \cite{trajectory2}. \textbf{2) Regime inference:} We measure the quality of the inferred regime assignments using Macro-F1 \citet{macrof1}. We report test results, with training and validation scores are provided in Appendix \ref{app:std}.

\textbf{Datasets} To assess HERON's ability to capture dynamical regimes, relational structure, and feature evolution, we evaluate it on both synthetic and real-world multi-agent datasets. The synthetic datasets cover three controlled settings: linear dynamics, which removes interactions to isolate temporal modeling and regime recovery; Gompertz dynamics \citep{gompertz1,gompertz2}, which introduce non-linear feature evolution over an evolving topology; and network interactions \citet{networkdynamic}, which focus on interaction-driven processes to test regime recovery from relational structure. The real-world datasets, SportVU NBA \citep{nba1,nba2} and AMASS \citep{amass} with BABEL annotations \citep{babel}, evaluate regime-aware forecasting in more complex settings. SportVU captures multi-agent trajectories with stable collective regimes defined by play outcomes, while for AMASS, we pool the original annotations into a coarser set of dynamical regimes Implementation details are provided in Appendix~\ref{app:dataset}, and dataset properties summarized in Table~\ref{tab:datasummary}.

\section{Forecasting performance with regime conditioning} \label{sec:result_forecasts}

We first evaluate how regime conditioning supports autonomous trajectory forecasting. Table~\ref{tab:all_results_simplified} shows that HERON remains competitive across all datasets and achieves the best ADE on the synthetic benchmarks, which cover complementary challenges: independent trajectories under a shared regime, non-linear dynamics with evolving topology, and interaction-driven network processes. In particular, strong performance on the network benchmark, where regimes are defined by relational propagation, indicates that HERON captures structured dynamics beyond simple temporal extrapolation.

\begin{table*}[ht]
\centering
\small
\begin{tabular}{lccccc}
\toprule
 & \multicolumn{3}{c}{Synthetic} & \multicolumn{2}{c}{Real-world} \\
\cmidrule(lr){2-4} \cmidrule(lr){5-6}
Model 
& Linear & Gompertz & Network & NBA & AMASS \\
\midrule
Persistence (Static)   
& 5.595 & 2.521 & 0.105 & 4.48/7.86 & 0.093/0.135 \\
Persistence (Velocity) 
& 2.896 & 2.906 & 0.058 & 2.55/5.75 & 0.129/0.271 \\
\midrule
S4 \citet{s4} 
& 4.072 & 4.307 & 0.162 & 2.09/5.11 & 0.422/0.616 \\
rSLDS \citet{rslds} 
& 2.744 & 3.171 & 0.052 & 2.93/5.40 & 0.074/0.144 \\
RED-SDS \citet{redsds} 
& 3.287 & 3.158 & 0.051 & 4.84/8.57 & 0.077/0.120 \\
TESTAM \citet{testam} 
& 2.975 & 2.661 & \underline{0.042} & 2.70/4.52 & \underline{0.055}/0.094 \\
EvolveGCN \citet{ref_evolve_gcn} 
& 3.454 & 2.623 & 0.057 & 1.72/4.92 & 0.060/0.088 \\
LaM-SLidE \citet{lam-slide} 
& \underline{2.077} & N/A & 0.054 & \textbf{0.73}/\textbf{1.32} & 0.061/0.084 \\
\midrule
GraphSSM \citet{graphssm} 
& 2.537 & 2.178 & 0.049 & 1.65/4.59 & 0.056/0.082 \\
GraphSSM+HERON \citet{graphssm} 
& 2.494 & \underline{2.042} & 0.047 & 1.59/4.43 & \underline{0.055}/\underline{0.080} \\
\midrule
HERON 
& \textbf{1.937} & \textbf{1.938} & \textbf{0.038} 
& \underline{0.81}/\underline{1.49} 
& \textbf{0.047}/\textbf{0.066} \\
\bottomrule
\end{tabular}
\caption{\textbf{Regime-conditioned forecasting performance.} 
Test ADE is reported for generated datasets, and ADE/FDE for motion datasets. 
Best results are in bold and second-best underlined. N/A means that the method was not applicable.}
\label{tab:all_results_simplified}
\end{table*}

On real-world datasets, HERON shows the intended forecasting--interpretability trade-off. On NBA, it remains close to LaM-SLidE, a strong latent trajectory model specialized for forecasting, while exposing an explicit system-level regime that can be tracked and evaluated. On AMASS, where motion regimes are fine-grained and may change within a sequence, HERON obtains the strongest ADE/FDE among the evaluated models. 

Furthermore, capacity-matched ablations show that gains are not explained by expert count alone. As shown in section \ref{sec:component_ablations}, removing regime conditioning or using fixed/random routing degrades forecasting and regime recovery, while removing the structural losses primarily weakens recovery. Overall, these results show that HERON preserves strong rollout accuracy while providing an explicit latent variable for modeling and analyzing dynamical regimes.

\section{HERON recovers meaningful regime} \label{sec:result_regime}
\paragraph{Recovery performance}
We evaluate whether the discrete states learned by HERON correspond to meaningful dynamical regimes rather than arbitrary latent clusters. Regime labels are not used during training and are only introduced for evaluation. Table~\ref{tab:regime-metric} shows strong alignment with ground-truth modes across synthetic datasets and NBA trajectories, supporting the role of the discrete state as an interpretable system-level variable. The network benchmark provides the most direct test of this claim. In this setting, regimes are defined by interaction mechanisms, rather than isolated node trajectories. HERON nevertheless recovers these modes with high Macro-F1 and AUC, indicating that the learned regime captures how information propagates through the graph. Together with the forecasting results, this shows that the regime variable is both predictive and aligned with the underlying dynamical mechanism.

\begin{table}[ht]
\centering
\small
\begin{tabular}{lccccc}
\toprule
 & \multicolumn{3}{c}{Synthetic} & \multicolumn{2}{c}{Real-world} \\
\cmidrule(lr){2-4} \cmidrule(lr){5-6}
Method & Linear & Gompertz & Network & NBA & AMASS \\
\midrule
Latent-HMM
& 0.945 & 0.897 & \underline{0.657} & \underline{0.593} & \underline{0.318} \\
rSLDS \cite{rslds}
& \underline{0.953} & 0.860 & 0.357 & 0.358 & 0.294 \\
RED-SDS \citet{redsds}
& 0.947 & \underline{0.926} & 0.557 & 0.451 & 0.299 \\
\midrule
HERON
& \textbf{0.961} & \textbf{0.927} & \textbf{0.913} 
& \textbf{0.891} & \textbf{0.412} \\
\bottomrule
\vspace{0.8pt}
\end{tabular}
\caption{\textbf{Self-supervised regime recovery.}
Test macro-F1 quantifies how well the learned regimes align with ground-truth dynamical labels, which are used only for evaluation.
Best results are in bold and second-best results are underlined.}
\label{tab:regime-metric}
\end{table}

The real-world datasets further evaluate fixed and time-varying regimes. On NBA, where regimes remain stable within each sequence, the learned states recover meaningful play patterns: the high macro-F1 shows that the model distinguishes rebound from scoring scenarios, which induce distinct motion patterns such as counterattacks and throw-ins. AMASS presents a more challenging setting, as action regimes may change within a sequence and the BABEL annotations provide only coarse proxies for the underlying motion dynamics reflected in the joint trajectories. Our extended analysis shows that  38\% of regime errors occur within five frames of a label transition, while macro-F1 increases from 0.412 to 0.566 on stable segments.  This suggests that smooth transitions and mismatch between semantic lables and trajectory dynamics account for part, but not all, of the remaining errors (Appendix~\ref{app:amass_errors}). Nevertheless, the learned states support strong forecasting and produce coherent regime-conditioned rollouts.

\paragraph{Interventional analysis of regimes}

\begin{wrapfigure}[23]{r}{0.67\textwidth}
\centering
\includegraphics[width=\linewidth]{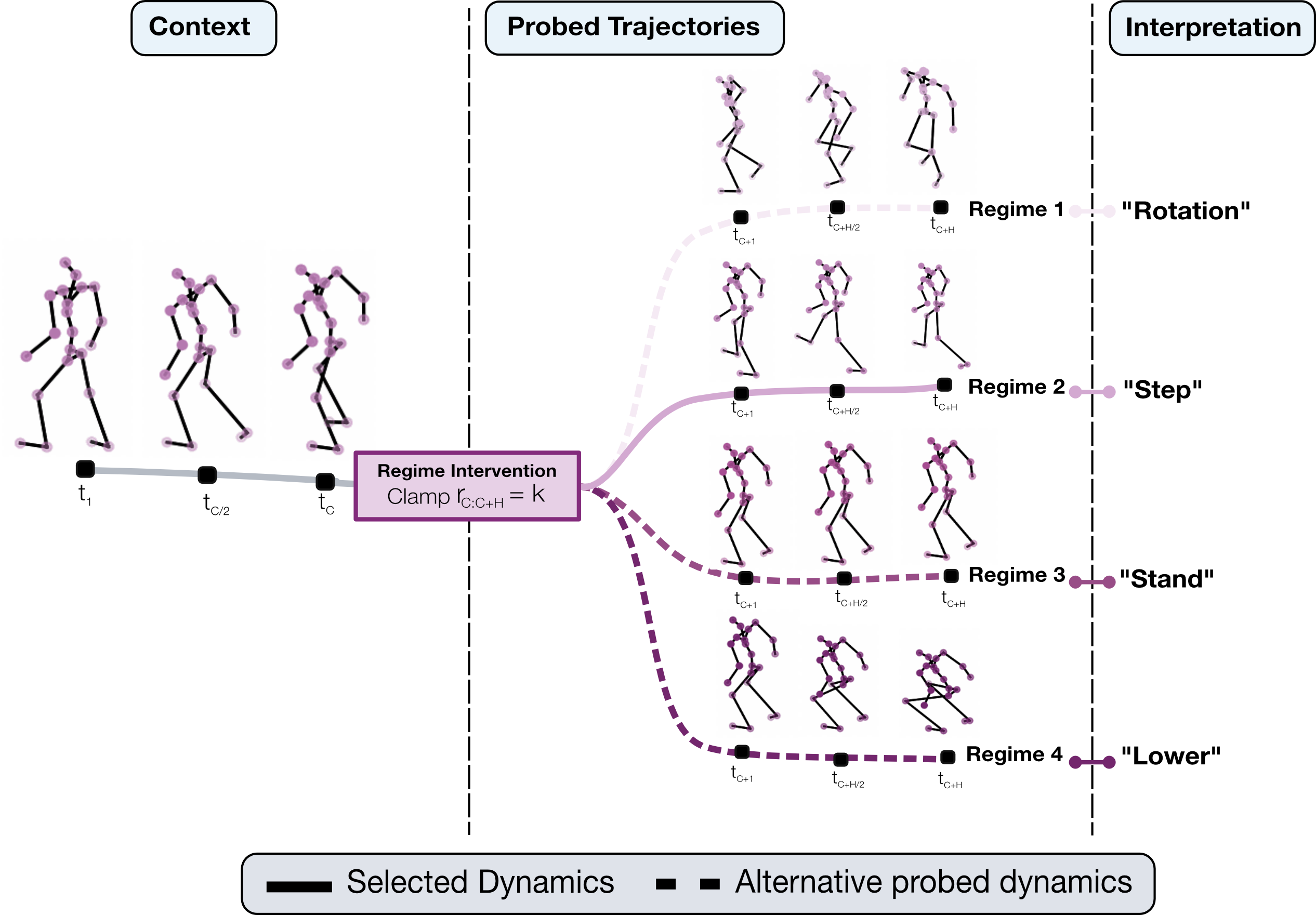}
\caption{\textbf{Regime-conditioned rollout probes.}
From a shared context, we clamp regimes during rollout. Dashed lines show regime-specific trajectories; the solid line is the observed future. Regime names are assigned post hoc from dominant motion patterns.}
\label{fig:amass}
\end{wrapfigure}

Finally, we probe the functional role of the regimes by intervening on the generator. Given the same context and recurrent state, we clamp the regime assignment to different experts and compare the resulting rollouts. Figure~\ref{fig:amass} shows that cchanging the clamped regime produces qualitatively distinct and recognizable motion patterns, such as walk or turn motions, rather than perturbing a single trajectory. Quantitatively, the matched regime yields the lowest ADE across datasets: clamping to an alternative regime increases ADE by 15--83\% on Network and 16--23\% on NBA, while physical and kinematic validity remain comparable under alternative clamps \ref{app:interventions}. These results support the central interpretation of HERON: the learned regime is not only recoverable from observed data, but also acts as an explicit conditioning variable for generating and probing alternative dynamical modes.

\section{Component ablations} \label{sec:component_ablations}
To determine whether HERON's performance gains arise from regime-aware
modeling rather than additional capacity, we ablate the regime variable,
its role in conditioning generation, the learned regime assignment, and
the structural objectives. We evaluate both forecasting and regime
recovery performances, and Table ~\ref{tab:component-ablations} reports test-set results under the same evaluation protocol as the main experiments.

The capacity-matched no-regime variant (A1) retains the same number and sizes of expert networks but averages their outputs without a discrete routing variable. Its degraded performance indicates that the gains cannot be explained by parameter count alone. When regimes are still inferred but no longer condition generation (A2), recovery remains substantial, while forecasting performance declines considerably. This contrast highlights the role of the inferred regime in governing prediction rather than merely providing descriptive labels. Consistently, replacing the learned assignment with fixed (A3a) or random routing (A3b) degrades both metrics, confirming the importance of context-dependent regime selection. Finally, removing the structural losses (A4) primarily weakens regime recovery but also reduces forecasting accuracy, indicating that these objectives help establish distinct and predictive regime representations. Overall, these results show that HERON's gains arise from learning context-dependent regimes and using them to govern generation, linking regime discovery to the dynamics the model predicts.

\begin{table*}[t]
\centering
\scriptsize
\setlength{\tabcolsep}{4pt}
\begin{tabular}{lccc|ccc}
\toprule
& \multicolumn{3}{c|}{Forecasting}
& \multicolumn{3}{c}{Regime recovery (Macro-F1)} \\
\cmidrule(lr){2-4} \cmidrule(lr){5-7}
Model
& Network ADE
& NBA ADE/FDE
& AMASS ADE/FDE
& Network
& NBA
& AMASS \\
\midrule
\textbf{HERON}
& \textbf{0.038}
& \textbf{0.81}/\textbf{1.49}
& \textbf{0.047}/\textbf{0.066}
& \textbf{0.913}
& \textbf{0.891}
& \textbf{0.412} \\

A1: no-regime, capacity-matched
& 0.048
& 1.52/2.97
& 0.055/0.078
& 0.248
& 0.546
& 0.290 \\

A2: passive regime
& 0.049
& 1.52/2.99
& 0.056/0.084
& \underline{0.786}
& \underline{0.782}
& \underline{0.369} \\

A3a: fixed regime
& 0.052
& 1.35/2.09
& \underline{0.053}/0.075
& 0.163
& 0.357
& 0.217 \\

A3b: random regime
& 0.049
& 1.38/2.09
& 0.055/0.075
& 0.371
& 0.501
& 0.307 \\

A4: no structural loss
& \underline{0.045}
& \underline{1.12}/\underline{1.83}
& 0.054/\underline{0.068}
& 0.557
& 0.527
& 0.317 \\
\bottomrule
\end{tabular}
\caption{\textbf{Component ablations.}
Test forecasting performance (ADE on Network and ADE/FDE on NBA and AMASS)
and test regime-recovery performance (Macro-F1). Lower forecasting errors
and higher Macro-F1 are better. Best results are in bold and second-best
results are underlined.}
\label{tab:component-ablations}
\end{table*}

\section{Prior work}

\textbf{Temporal graph methods.} Temporal graph models extend GNNs to evolving relational systems, where node features, edges, or topology change over time. Recurrent and evolution-based GNNs update node embeddings or graph parameters across snapshots \citep{ref_evolve_gcn,gnnrecurrent1}, while continuous-time temporal graph networks maintain event-driven memories \citep{tgn1,tgn2,tgn3}. Spatio-temporal GNNs combine relational message passing with temporal modeling for structured forecasting \citep{stgnn1,stgnn2,stgnn3}, and recent graph state-space models further integrate relational mixing within latent dynamical systems \citep{graphssm,gssm1}. However, these representations are primarily optimized for prediction, with spatial aggregation, temporal memory, and rollout dynamics tightly coupled, making the system-level dynamical mode difficult to identify.

\textbf{Latent dynamical models.} Latent dynamical and neural state-space models represent trajectories through hidden states and transition dynamics. Classical and neural formulations learn compact latent dynamics for sequential prediction \citep{ogssm1,ogssm2,ssm1}, while structured models such as S4 \citet{s4} and Mamba \cite{mamba} scale these ideas to long-range dependencies, and world models support autonomous latent rollouts and planning \citep{planet,dreamerv2,dreamerv3,lam-slide}. These models achieve strong predictive performance, but their latent states are generally not exposed as interpretable regimes that directly govern generation. Conversely, switching dynamical systems \citep{regimeinference1,regimeinference2} and change-point models \citep{changepoint1,changepoint2,changepoint3} explicitly represent discrete modes, but typically focus on identifying them in observed sequences rather than generating future graph-structured trajectories.

HERON bridges this gap by unifying graph-structured trajectory generation with self-supervised regime identification, enabling forecasts that are predictable and regimes that are actionable.

\section{Conclusion and future directions}
\label{sec:conclusion}

We introduced HERON, a self-supervised hierarchical framework that unifies regime inference and spatio-temporal feature generation on graphs through discrete system-level regimes that condition entity-level trajectory generation. Across controlled and real-world datasets, HERON recovers meaningful regimes while remaining competitive with strong forecasting baselines. Moreover, clamping the regime during rollout from a shared context produces distinct yet coherent alternative trajectories. These results support the central premise of this work: regimes are not merely post-hoc explanations but explicit conditioning variables that enable forecasting, tracking, and controlled simulation of alternative dynamical modes. Adaptive regime discovery and modeling topological events, such as node appearance and disappearance, as part of the generative process remain promising directions (see Appendix ~\ref{app:limitations} for a more detailed discussion). Furthermore, extending HERON across different relational architectures and data modalities represents an important direction for future work, particularly in domains such as biomedicine, where regime inference could reveal meaningful latent states associated with disease progression or patient dynamics.

\section{Acknowledgements}
    This work was partially supported by the Swiss National Science Foundation (SNSF) under grant IZVSZ2\_229539 and the PHRT-SDSC project under grant 2022-811/C22-16P. We thank the anonymous reviewers for their insightful feedback and constructive discussion.

\bibliography{DigitalPathology.bib}

\clearpage

\tableofcontents

\section{Impact Statement}
\label{broader_impacts}
HERON aims to improve the forecasting and understanding of evolving relational systems by identifying the dynamical regimes governing their evolution. This approach could support more informed decision-making in sensitive domains such as healthcare, epidemiology, and finance. While we do not currently foresee any immediate societal concerns arising from the proposed methodology, inaccurate regime identification or forecasts could lead to misleading conclusions. Moreover, identified regimes should not be interpreted as causal explanations without further validation. We therefore emphasize the importance of rigorous evaluation and human oversight when deploying HERON in sensitive domains.

\section{Appendix} \label{sec:appendix}

This appendix provides the implementation and evaluation details that are most relevant for assessing HERON as a regime-aware latent modeling framework. The main text presents the conceptual model and empirical results; here we clarify how the method can be instantiated on top of existing latent graph-temporal models, how posterior and prior paths differ, how the generated datasets are constructed, and how evaluation is performed.

\subsection{HERON}

\subsubsection{Hierarchical states modeling} \label{app:latent_comp}

\textbf{Latent state initialization.}

We initialize the hierarchical latent state as follows. For each entity $i$, let $\tau_i$ denote the first time step at which the entity is observed. The entity-level latent state and deterministic recurrent state are initialized as
\[
z_{\tau_i}^{i} = \mathbf{0},
\qquad
h_{\tau_i}^{i} = \mathbf{0}.
\]
Thus, entities present at the beginning of the sequence are initialized at the first time step, while entities entering later are initialized when they first appear.

At the system level, the deterministic recurrent state is initialized analogously:
\[
h_{0}^{G} = \mathbf{0}.
\]
The initial regime distribution is set to the uniform categorical prior over the $K$ regimes,
\[
p(r_0 = k) = \frac{1}{K},
\qquad k \in \{1,\ldots,K\},
\]
or equivalently,
\[
\boldsymbol{\pi}_0 = \frac{1}{K}\mathbf{1}_K .
\]
This initialization encodes a non-informative prior over regimes at the first time step, assigning equal probability to each regime. Alternative initial regime priors can be incorporated in the same manner as the load-balancing procedure described in Appendix~\ref{app:loss}.

\textbf{Regime discrete variable.}

For a graph snapshot index by $b$ in the batch, the system-level transition module produces regime logits $\ell_b$. During training, we use the Gumbel--Softmax reparameterization to sample regimes while retaining gradient information.
First, we generate Gumbel noise from uniform random variables:
\begin{equation}
    g_{b,k} = -\log(-\log u_{b,k}),
    \qquad u_{b,k} \sim \mathrm{U}(0,1).
    \tag{8}
\end{equation}

We then add this noise to the regime logits and apply a
temperature-scaled softmax to obtain a differentiable, relaxed
regime assignment:
\begin{equation}
    \tilde{r}_{b,k}
    =
    \frac{\exp((\ell_{b,k}+g_{b,k})/\tau)}
    {\sum_{j=1}^{K}\exp((\ell_{b,j}+g_{b,j})/\tau)},
    \qquad k \in \{1,\ldots,K\}.
    \tag{9}
\end{equation}

The temperature $\tau$ is annealed during training, typically from
$1.0$ to $0.3$ over 60\% of training, encouraging exploration early
and sharper assignments later. Finally, we obtain a one-hot regime
assignment using the argmax, while the straight-through estimator
preserves gradient flow through the relaxed assignment:
\begin{equation}
    r_{b,k}^{\mathrm{hard}}
    =
    \mathbb{1}\left[k=\arg\max_j \tilde{r}_{b,j}\right],
    \qquad
    r_b
    =
    r_b^{\mathrm{hard}}
    - \operatorname{sg}(\tilde{r}_b)
    + \tilde{r}_b.
    \tag{10}
\end{equation}

Here, $\ell_{b,k}$ denotes the logit for regime $k$ in graph $b$,
and $g_{b,k}$ denotes the corresponding Gumbel noise. The relaxed
assignment $\tilde{r}_b$ supports gradient propagation, while
$r_b^{\mathrm{hard}}$ provides the one-hot assignment used in the
forward pass.

\textbf{Entity-level latent.} The entity states are modeled as Gaussian stochastic variables. For each entity $i$, the posterior and prior transition heads output the parameters of a diagonal Gaussian distribution:
\begin{equation}
q_{\phi}(z_t^i \mid h_t^i, \mathrm{emb}_t^i, r_t)
=
\mathcal{N}\!\left(
\mu_{\phi,t}^i,
\operatorname{diag}\!\left((\sigma_{\phi,t}^i)^2\right)
\right),
\end{equation}
with an analogous parameterization for the prior distribution defined as
\begin{equation}
p_{\psi}(\hat z_t^i \mid h_t^i, \widehat{\mathrm{emb}}_t^i, \hat r_t)
=
\mathcal{N}\!\left(
\mu_{\psi,t}^i,
\operatorname{diag}\!\left((\sigma_{\psi,t}^i)^2\right)
\right).
\end{equation}
The regime variable therefore provides a discrete system-level assignment, while the Gaussian entity latents represent continuous node-level uncertainty conditioned on the active regime.

\subsubsection{Mixture-of-experts as a transition model.}\label{app:moe}

Our formulation in Section~\ref{sec:problem_formulation} assumes that the modeled system evolves according to $K$ distinct transition functions $\{f_k\}_{k=1}^K$, each corresponding to a different dynamical regime. This assumption naturally motivates the use of Mixture-of-Experts models~\citep{moe1,moe2}, since they explicitly parameterize a collection of candidate transition functions and select or combine them through a gating mechanism. In our implementation, we use a one-hot regime assignment rather than a dense mixture. This design choice encourages a clearer separation between regimes and reduces the risk that multiple experts jointly encode indistinguishable dynamics.

The MoE parameterization also provides a direct mechanism for controlled interventions in latent rollouts. At generation time, the selected expert can be clamped, allowing the model to simulate trajectories under a fixed regime rather than allowing the gating network to select regimes autonomously. This enables controlled visualization of alternative plausible futures and is a direct consequence of the explicit regime-conditioned transition structure.

\subsubsection{Posterior and Prior Inference Paths} \label{app:posterior_prior}

HERON maintains two inference paths. The \emph{posterior} path is used when the current graph observation is available; the \emph{prior} path is used during rollout. Variables without hats denote posterior quantities, and variables with hats denote prior quantities. Unless explicitly stated, parameters indexed by $\phi$ and $\psi$ are separate modules with no shared weights. In our implementation, the deterministic recurrent updates and the decoder are shared across both paths, while the posterior and prior embedding networks and stochastic transition heads are distinct. Algorithm \ref{alg:HERON} show this hierarchical process in details.

\begin{algorithm}[t]
\caption{HERON hierarchical inference and rollout}
\label{alg:HERON}
\begin{algorithmic}[1]
\Require Context sequence $(\mathbf{X}_{1:C},\mathcal{V}_{1:C},\mathcal{E}_{1:C})$, target topology $(\mathcal{V}^\star,\mathcal{E}^\star)$, context size $C$, horizon $H$
\Ensure Forecasts $\hat{\mathbf{X}}_{C+1:C+H}$ and regime trajectory $r_{1:C+H}$
\State $T \gets C + H$
\State Initialize $h_0^G,\{h_0^i\}_{i \in \mathcal{V}_1},r_0,\{z_0^i\}_{i \in \mathcal{V}_1}$
\For{$t=1$ to $T$}
    \State ${h}^i_t \gets GRU_\theta^\mathcal{V}({h}^i_{t-1},{z}^i_{t-1})$, \quad $h_t^G \gets GRU_\theta^G(h_{t-1}^G,r_{t-1})$ \Comment{\textit{Recurrent state update}}
    \If{$t \leq C$}
        \State $\mathrm{emb}^i_t \gets \mathrm{GNN}_\phi({h}^i_t,\mathbf{X}_t,{\mathcal{E}}_t)$,
        \quad $\mathrm{emb}_t^G \gets \mathrm{Pool}_\phi(\{\mathrm{emb}^i_t\},{\mathcal{E}}_t)$ \Comment{\textit{Posterior embeddings}}
        \State $r_t \sim q_\phi(r_t \mid h_t^G,\mathrm{emb}_t^G)$, \quad ${z}^i_t \sim q_\phi({z}^i_t \mid {h}^i_t,\mathrm{emb}^i_t,r_t)$ \Comment{\textit{Posterior state inference}}
        \State $\hat{\mathbf{X}}_{t+1} \sim g_\theta(\mathbf{X}_{t+1}\mid h_t^G,\{h_t^i\}_{i \in \tilde{\mathcal{V}}_t},r_t,\{z_t^i\}_{i \in \tilde{\mathcal{V}}_t})$ \Comment{\textit{Posterior decoding}}
    \Else
        \State $\hat{\mathrm{emb}}^i_t \gets \mathrm{GNN}_\psi({h}^i_t,{\mathcal{E}}^\star)$,
        \quad $\hat{\mathrm{emb}}_t^G \gets \mathrm{Pool}_\psi(\{\hat{\mathrm{emb}}^i_t\},{\mathcal{E}}^\star)$ \Comment{\textit{Prior embeddings}}
        \State $\hat r_t \sim p_\psi(\hat r_t \mid h_t^G,\hat{\mathrm{emb}}_t^G)$, \quad ${\hat z}^i_t \sim p_\psi({\hat z}^i_t \mid {h}^i_t,\hat{\mathrm{emb}}^i_t,\hat r_t)$ \Comment{\textit{Prior state inference}}
        \State $\hat{\mathbf{X}}_{t+1} \sim g_\theta(\mathbf{X}_{t+1}\mid h_t^G,\{h_t^i\}_{i \in \tilde{\mathcal{V}}_t},\hat r_t,\{\hat z_t^i\}_{i \in \tilde{\mathcal{V}}_t})$ \Comment{\textit{Prior decoding}}
    \EndIf
\EndFor
\end{algorithmic}
\end{algorithm}

\paragraph{Posterior path.}

Starting from the updated deterministic states $h_t$, the posterior path then uses the current observed node features and graph topology to form relational embeddings:
\begin{equation}
\mathrm{emb}_t^i
=
\mathrm{GNN}_{\phi}\!\left(h_t^i,\mathbf{X}_t,\mathcal{E}_t\right),
\qquad
\mathrm{emb}_t^G
=
\mathrm{Pool}_{\phi}\!\left(\{\mathrm{emb}_t^i\}_{i\in\mathcal{V}_t},\mathcal{E}_t\right).
\end{equation}
The graph-level embedding produces the categorical regime posterior:
\begin{equation}
r_t \sim q_{\phi}\!\left(r_t \mid h_t^G,\mathrm{emb}_t^G\right).
\end{equation}
The sampled regime then conditions the entity-level latent posterior:
\begin{equation}
z_t^i \sim q_{\phi}\!\left(z_t^i \mid h_t^i,\mathrm{emb}_t^i,r_t\right),
\qquad i\in\mathcal{V}_t.
\end{equation}
This path is observation-conditioned. It is used to infer latent states from the context and to provide a training target for the prior.

\paragraph{Prior path.}
Starting from the updated deterministic states $h_t$, the prior path embeddings are computed from recurrent states and topology alone:
\begin{equation}
\hat{\mathrm{emb}}_t^i
=
\mathrm{GNN}_{\psi}\!\left(h_t^i,\hat{\mathcal{E}}_t\right),
\qquad
\hat{\mathrm{emb}}_t^G
=
\mathrm{Pool}_{\psi}\!\left(\{\hat{\mathrm{emb}}_t^i\}_{i\in\hat{\mathcal{V}}_t},\hat{\mathcal{E}}_t\right).
\end{equation}
The prior regime is sampled from the graph-level prior:
\begin{equation}
\hat r_t \sim p_{\psi}\!\left(\hat r_t \mid h_t^G,\hat{\mathrm{emb}}_t^G\right).
\end{equation}
Finally, the regime-conditioned node-level prior produces entity latent states:
\begin{equation}
\hat z_t^i
\sim
p_{\psi}\!\left(\hat z_t^i \mid h_t^i,\hat{\mathrm{emb}}_t^i,\hat r_t\right),
\qquad i\in\hat{\mathcal{V}}_t.
\end{equation}

\subsubsection{Loss Implementations}
\label{app:loss}

\paragraph{Prediction loss.}
To encourage the latent regime to capture dynamical evolution rather than static feature configurations, the decoder is trained to predict node-level feature velocities,
\begin{equation}
\mathbf{Y}_t = \mathbf{X}_{t+1} - \mathbf{X}_t .
\end{equation}
This velocity target is used only as the training target. During inference, predictions are accumulated and evaluated in the original feature space. The decoder formulation in the main text is kept in terms of feature predictions for consistency with the rollout objective and evaluation protocol. Note that prediction loss is only applied to time-varying features. Any static node or edge features will be seen by the model as input but will not be reconstructed.

During the context window, the prediction loss is applied to both posterior and prior paths. This parallel decoding reduces the mismatch between the observation-conditioned states used during context and the autonomous prior states used after the context window. During the forecast horizon, the loss is applied only to the prior path, since future observations are unavailable and the posterior path cannot be instantiated.

\paragraph{Balanced KL losses.}
The KL terms are computed analytically for both components of the hierarchical stochastic state. For the discrete regime variable, the KL divergence is evaluated between categorical distributions over the $K$ regimes as follows:

\begin{equation}
\mathrm{KL}\!\left[
q_{\phi}(r_t) \,\|\, p_{\psi}(\hat r_t)
\right]
=
\sum_{k=1}^{K}
q_{\phi}(r_t=k)
\log
\frac{q_{\phi}(r_t=k)}
     {p_{\psi}(\hat r_t=k)} .
\end{equation}

For the entity-level latent states, the KL divergence is evaluated in closed form between diagonal Gaussian distributions: 

\begin{equation}
\mathrm{KL}\!\left[
q_{\phi}(z_t^i) \,\|\, p_{\psi}(\hat z_t^i)
\right]
=
\frac{1}{2}
\sum_{d=1}^{d_z}
\left[
\log
\frac{(\sigma_{\psi,t,d}^{i})^2}
     {(\sigma_{\phi,t,d}^{i})^2}
+
\frac{
(\sigma_{\phi,t,d}^{i})^2
+
(\mu_{\phi,t,d}^{i}-\mu_{\psi,t,d}^{i})^2
}
{(\sigma_{\psi,t,d}^{i})^2}
-1
\right].
\end{equation}

These terms are used in the balanced KL formulation described in the main text, where the dynamics term $\mathcal{L}_{\mathrm{dyn}}
$ trains the prior to match the stopped-gradient posterior and the representation term $\mathcal{L}_{\mathrm{rep}}$ regularizes the posterior toward the stopped-gradient prior. We define the two losses as the following:

\begin{align}
\mathcal{L}_{\mathrm{dyn}}
&=
\mathrm{KL}\!\left[
\operatorname{sg}\!\left(q_{\phi}(r_t)\right)
\,\|\,
p_{\psi}(\hat r_t)
\right]
+
\frac{1}{|\mathcal{V}_t|}\sum_{i\in \mathcal{V}_t}
\mathrm{KL}\!\left[
\operatorname{sg}\!\left(q_{\phi}(z_t^i)\right)
\,\|\,
p_{\psi}(\hat z_t^i)
\right],
\\
\mathcal{L}_{\mathrm{rep}}
&=
\mathrm{KL}\!\left[
q_{\phi}(r_t)
\,\|\,
\operatorname{sg}\!\left(p_{\psi}(\hat r_t)\right)
\right]
+
\frac{1}{|\mathcal{V}_t|}\sum_{i\in \mathcal{V}_t}
\mathrm{KL}\!\left[
q_{\phi}(z_t^i)
\,\|\,
\operatorname{sg}\!\left(p_{\psi}(\hat z_t^i)\right)
\right].
\end{align}

\paragraph{Structural loss.}
The structural regularizers are applied primarily to the posterior path. The balanced KL objective then transfers these constraints to the prior by aligning the prior latent distributions with the posterior ones. This design avoids imposing all structural constraints directly on the autonomous rollout path while still encouraging the prior to reproduce structured regime assignments.

The \textbf{load-balancing} term encourages the model to use its
$K$ experts in a balanced manner across the batch. We first obtain
the regime probabilities by applying a softmax to the regime logits
without Gumbel noise. For graph with index $b$ in the batch and regime $k$, this gives
\begin{equation}
    \pi_{b,k}
    =
    \operatorname{softmax}(\ell_b)_k.
\end{equation}

We then average these probabilities over the $B$ graphs in the batch
to estimate the empirical usage of each regime, denoted by
$\bar{\pi}_k$. We compare this distribution against a uniform target
$u$, which assigns equal probability to each of the $K$ regimes:
\begin{equation}
    \bar{\pi}_k
    =
    \frac{1}{B}
    \sum_{b=1}^{B} \pi_{b,k},
    \qquad
    u_k = \frac{1}{K}.
\end{equation}

The load-balancing loss is defined as the KL divergence between the
empirical regime distribution and this uniform target. Substituting
$u_k = 1/K$ gives the following equivalent expression:
\begin{equation}
    \begin{aligned}
    \mathcal{L}_{\mathrm{balancing}}
    &=
    \mathrm{KL}\!\left[
        \bar{\pi} \,\|\, u
    \right] \\
    &=
    \sum_{k=1}^{K}
        \bar{\pi}_k
        \log \frac{\bar{\pi}_k}{u_k} \\
    &=
    \sum_{k=1}^{K}
        \bar{\pi}_k
        \log\left(K\bar{\pi}_k\right).
    \end{aligned}
\end{equation}

The uniform target reflects the assumption that regimes are
approximately balanced in the datasets considered in this work.
Under severe regime imbalance, however, this assumption becomes
restrictive. In Appendix~\ref{app:imbalance}, we therefore describe
an extension in which the regime prior is learned from the data
rather than fixed to be uniform. In a controlled setting with a
70\%/20\%/10\% regime distribution, enforcing the uniform prior
causes minority-regime recall to fall to zero. Replacing the fixed
uniform target with an unsupervised learned categorical prior
recovers minority-regime recall to 0.70 and improves Macro-F1
from 0.60 to 0.78.

The \textbf{entropy regularizer} is computed on the sample-level regime probabilities before Gumbel perturbation. Minimizing this entropy encourages low-entropy regime assignments, making the inferred regime distribution closer to one-hot and improving the confidence of expert selection:

\begin{equation}
\mathcal{L}_{\mathrm{entropy}} = \mathcal{H}\!\left[Q_t\right]
=
\frac{1}{B}
\sum_{b=1}^{B}
H(\pi_{b,t})
=
-
\frac{1}{B}
\sum_{b=1}^{B}
\sum_{k=1}^{K}
\pi_{b,t,k}
\log \pi_{b,t,k}.
\end{equation}

Finally, the \textbf{expert-specialization} term is computed using all expert outputs, even though the effective gating selects a single expert through the straight-through regime assignment. The loss is applied to the mean component of the Gaussian entity latent distribution, rather than to sampled latent vectors, to reduce sensitivity to sampling noise. This penalizes alignment between different expert outputs for the same entity and encourages regime-specific latent mappings. The loss can formally be writen as

\begin{equation}
\mathcal{L}_{\mathrm{ortho}}
=
\sum_{i\in V_t}
\sum_{j=1}^{K}
\sum_{\substack{k=1 \\ k\neq j}}^{K}
\left\|
\frac{
\left\langle
\mu_{\phi,j}^{i},
\mu_{\phi,k}^{i}
\right\rangle
}
{
\left\langle
\mu_{\phi,k}^{i},
\mu_{\phi,k}^{i}
\right\rangle
+\epsilon
}
\mu_{\phi,k}^{i}
\right\|_2^2 ,
\end{equation}

where $\langle \cdot \rangle$ denotes the inner product between two vectors, and $\mu_{\phi,k}^{i}$ represents the mean component of the Gaussian output of expert $k$ for entity $i$ after the MoE (see section \ref{app:latent_comp} for mode detail on notation). We found this latent-space formulation preferable to weight-space orthogonality alternatives, which were less effective in our experiments. Future work may consider lower-cost specialization mechanisms, for example by parameterizing experts as basis functions, although such restrictions may reduce expressivity.

\subsection{Experimental Protocol and Metrics}
\label{app:protocol}

Sequences are split at the sequence level, where windows from the same original sequence are never allowed to appear in different splits. We use fixed windows of length $10$ for generated datasets, with $C=5$ context steps and $H=5$ rollout steps. For real-world datasets NBA \citep{nba1,nba2} and AMASS \citep{amass}, we use longer windows ($C=8,H=12)$; the context and horizon lengths are set according to the preprocessing protocol used for the corresponding dataset.

Hyperparameters are selected with four-fold cross-validation on the training split. All reported forecasting metrics are computed on the horizon using the prior rollout path, not the posterior path. This ensures that evaluation matches the intended deployment setting where future observations are unavailable.

Let $\mathcal{B}$ be the set of test windows, and let $\mathcal{V}_{b,t}$ be the nodes evaluated in window $b$ at time $t$. The Average Displacement Error is
\begin{equation}
\mathrm{ADE}
=
\frac{1}{|\mathcal{B}|H}
\sum_{b\in\mathcal{B}}
\sum_{t=C+1}^{C+H}
\frac{1}{|\mathcal{V}_{b,t}|}
\sum_{i\in\mathcal{V}_{b,t}}
\left\|
\hat x_{b,t}^i-x_{b,t}^i
\right\|_2 .
\end{equation}
The Final Displacement Error is
\begin{equation}
\mathrm{FDE}
=
\frac{1}{|\mathcal{B}|}
\sum_{b\in\mathcal{B}}
\frac{1}{|\mathcal{V}_{b,C+H}|}
\sum_{i\in\mathcal{V}_{b,C+H}}
\left\|
\hat x_{b,C+H}^i-x_{b,C+H}^i
\right\|_2 .
\end{equation}
More precisely, we report the minimum FDE across several stochastic rollouts samples ($M=20$):
\begin{equation}
\mathrm{minFDE}
=
\frac{1}{|\mathcal{B}|}
\sum_{b\in\mathcal{B}}
\min_{m\in\{1,\ldots,M\}}
\frac{1}{|\mathcal{V}_{b,C+H}|}
\sum_{i\in\mathcal{V}_{b,C+H}}
\left\|
\hat x_{b,C+H}^{i,(m)}-x_{b,C+H}^i
\right\|_2 .
\end{equation}

For regime evaluation, labels are used only after training. Because discrete latent states are identifiable only up to permutation, we first align predicted regime IDs with ground-truth labels using the Hungarian algorithm on the validation confusion matrix. Macro-F1 is then computed after this alignment. For AUC, we permute the predicted regime probabilities according to the same assignment and compute one-vs-rest AUC for multi-class settings.

\subsection{Baselines}
\label{app:baselines}

The baselines are chosen to contextualize forecasting quality, not to claim that all methods solve the same regime-inference problem. Persistence baselines test whether the task can be solved by short-term extrapolation. S4 evaluates a strong non-relational sequence model. EvolveGCN represents recurrent temporal graph modeling. GraphSSM and LaM-SLidE provide latent spatio-temporal forecasting references.

All baselines are evaluated under the same data splits, context length $C$, horizon $H$, autoregressive rollout protocol, and forecasting metrics as HERON. Learned baselines are tuned using the four-fold cross-validation protocol in Section~\ref{app:protocol}. For methods that do not expose an explicit regime variable, we report forecasting metrics only.

\paragraph{Persistence (Static and Velocity).}
We include two parameter-free temporal heuristics: Static, $\hat{X}_{t+1}=X_t$, and Velocity, $\hat{X}_{t+1}=2X_t-X_{t-1}$. Both are applied to the time-varying channels only, while static descriptors are copied forward unchanged. Rollouts are autoregressive, using previous predictions as inputs after the context window. For Velocity, the first predicted step uses the last observed difference $X_C-X_{C-1}$; newly appearing nodes in Gompertz default to Static at their entry step.

\paragraph{S4 \citet{s4}.}
S4 is used as a non-relational state-space baseline. It models temporal dependencies with structured state-space layers but does not use graph topology or edge features. Given node features $X \in \mathbb{R}^{B \times N \times C \times d_V}$, we apply shared S4 blocks independently to each node trajectory by reshaping the input as $(B N) \times C \times d_V$. To allow limited graph-level context without using edges, the resulting node states are aggregated with a masked mean over the entities present in each graph and injected back into each node representation through a linear residual followed by layer normalization:
\[
H = \mathrm{S4Block}\!\left(X.\mathrm{reshape}(BN,C,d_V)\right).\mathrm{reshape}(B,N,C,d_h),
\]
\[
c_{b,t} = \frac{1}{|\mathcal{V}_{b,t}|}\sum_{i \in \mathcal{V}_{b,t}} H_{b,i,t},
\qquad
\tilde{H}_{b,i,t} = \mathrm{LayerNorm}\!\left(H_{b,i,t} + W_c c_{b,t}\right).
\]
A regression head maps $\tilde{H}$ to the next-step target distribution, and horizon predictions are produced autoregressively.

\paragraph{rSLDS \citet{rslds}.}

rSLDS is the canonical classical switching baseline, and reports both forecasting and regime recovery. We use the Linderman lab \texttt{ssm} implementation in its recurrent-only variant, in which the discrete regime is driven by the continuous state rather than by its own past:
\[
p(z_t = k \mid x_{t-1}) \propto \exp\!\left(R_k^\top x_{t-1} + r_k\right),
\qquad
x_t = A_{z_t} x_{t-1} + b_{z_t} + \epsilon_t,
\qquad
y_t = C x_t + d + \eta_t ,
\]
fitted by Laplace-EM on the training fold. Being non-relational, the model needs a flat observation: NBA and AMASS frames are flattened node-major into a single velocity vector (22 and 66 dimensions), the latter reduced to 15 PCA components; on Network, where the node set differs across sequences, one shared model is fitted to all per-node trajectories. Forecasts are autonomous rollouts seeded from the posterior-filtered terminal context state, and regimes are read from the discrete posterior over the context.

\paragraph{RED-SDS \citet{redsds}.}

RED-SDS is the considered deep switching baseline. It augments the regime $z_t$ with a count variable $c_t$, so dwell times are modelled explicitly rather than implied by a self-transition probability. With $u_k(c)$, we define the probability of remaining in regime $k$ after $c$ steps. The joint regime-duration transition is then:
\[
\begin{aligned}
p(z_t, c_t \mid z_{t-1}, c_{t-1}, x_{t-1})
&= u_{z_{t-1}}(c_{t-1})\,\delta[z_t{=}z_{t-1}]\,\delta[c_t{=}c_{t-1}{+}1] \\
&\quad + \big(1 - u_{z_{t-1}}(c_{t-1})\big)\,
p(z_t \mid z_{t-1}, x_{t-1})\,\delta[c_t{=}1].
\end{aligned}
\]
over nonlinear transitions $x_t \sim \mathcal{N}(f_{z_t}(x_{t-1}), \cdot)$ and emissions $y_t \sim \mathcal{N}(g(x_t), \cdot)$, trained by maximising an ELBO with annealed relaxations of both $z$ and $c$. We use the authors' implementation unmodified, adding only a dataset adapter that reuses the flattening and PCA front-end of rSLDS \citet{rslds}, with the maximum duration set to the window length. Forecasts come from the open-loop unroll, regimes from the smoothed posterior marginals.

\paragraph{TESTAM \citet{testam}.}

TESTAM is the graph mixture-of-experts baseline: it routes implicitly and locally, where HERON commits to an explicit system-level regime. Three experts defined as temporal, adaptive-graph (supports generated from a learned memory), and spatial-attention are combined by a gate selecting one expert per node and time step,
\[
g = \mathrm{softmax}\big(\mathrm{Gate}(h^{\mathrm{tmp}}, h^{\mathrm{adp}}, h^{\mathrm{att}})\big),
\qquad
\hat{y} = o_{e^\star}, \quad e^\star = \arg\max_e g_e ,
\]
trained with the authors' pseudo-label routing losses, which push the gate away from the worst expert and towards the best. We use the model unchanged. As our windows carry no time-of-day signal, its time channel becomes a within-window positional index, and multi-channel node features are emitted through the expert output width, one routing decision being shared across channels. On Network, whose node set varies, sequences are zero-padded to a fixed node table. TESTAM is forecasting-only.

\paragraph{EvolveGCN \citet{ref_evolve_gcn}.}
EvolveGCN is a recurrent temporal graph baseline in which the GCN parameters evolve over time. We use the EvolveGCN-H variant, where the layer weights are updated by a GRU driven by the current node embeddings:
\begin{equation}
W_t^{(\ell)} = \mathrm{GRU}\!\left(H_t^{(\ell-1)},\, W_{t-1}^{(\ell)}\right),
\qquad
H_t^{(\ell)} = \sigma\!\left(\hat{A}_t H_t^{(\ell-1)} W_t^{(\ell)}\right).
\end{equation}
We attach a two-layer MLP regression head to the final node embeddings $H_t^{(L)}$ and train the model with the same negative Gaussian log-likelihood as HERON. The model is unrolled over the context and prediction windows, consuming observed features during the context and its own predictions during the horizon. We use a two-layer EvolveGCN-H stack, matching the depth of the graph encoder in HERON.

\paragraph{LaM-SLidE \citet{lam-slide}.}
LaM-SLidE is a latent generative model for spatial dynamical systems. It first trains an entity-aware autoencoder using learned entity identifiers, then trains a flow-matching prior over latent trajectories conditioned on a context window. We use the authors' public implementation with task-specific tokenization. NBA entities are represented by fixed player/ball identifiers together with coordinates and type indicators; AMASS joints are represented by joint identifiers and anatomical descriptors. The latent prior is trained to generate $H$ future latent steps from the $C$ context latents. At evaluation, we draw $M=20$ stochastic samples per window, matching the minFDE protocol in Section~\ref{sec:experiments}. We evaluate LaM-SLidE on the fixed-entity datasets only; synthetic datasets with variable entity sets  are marked as not applicable.

\paragraph{GraphSSM \citet{graphssm}.}
GraphSSM extends structured state-space models to temporal graphs by combining temporal SSM updates with graph mixing through a Laplacian-based GHiPPO formulation. We use GraphSSM-S4, the strongest variant reported by the original authors on temporal-graph benchmarks. For comparison with HERON, we stack the same number of GraphSSM-S4 blocks as the HERON encoder, replace the original task head with the same MLP velocity decoder, and train with the identical negative Gaussian log-likelihood. The Laplacian is recomputed at every snapshot for Gompertz, where topology changes over time, and once per window otherwise. GraphSSM-S4 serves as the closest non-regime-aware graph state-space comparator.

\paragraph{Latent-HMM.}

The HMM baseline adds memory to the simple regime-discovery controls, which cluster one pooled vector per sequence. A $K$-state Gaussian HMM with diagonal covariances is fitted by EM on per-timestep, node-pooled feature sequences $f_{1:T}$. Its likelihood can be written as:

$$p(f_{1:T}) = \sum_{s_{1:T}} \pi_{s_1} \prod_{t>1} A_{s_{t-1} s_t} \prod_t \mathcal{N}\left(f_t; \mu_{s_t}, \mathrm{diag},\sigma^2_{s_t}\right),$$

so neighbouring frames are tied by a learned transition law rather than treated independently. We use the GraphSSM latent as the velocity target, meaning the per-graph mean of the pre-head hidden state of a frozen GraphSSM trained only to forecast the next step. Features are standardised on the training fold.

\subsection{Datasets} \label{app:dataset}

\begin{table}[ht] 
\captionsetup{skip=10pt}
\centering
\small
\begin{tabular}{lccccc} 
\toprule
 & \multicolumn{3}{c}{Synthetic} & \multicolumn{2}{c}{Real-world} \\
\cmidrule(lr){2-4} \cmidrule(lr){5-6}
 & Linear & Gompertz & Network & NBA & AMASS \\
\midrule
Number of entities per graph   & 50-100 & 50-100 & 50-100 & 11 & 22 \\
Number of sequences   & $\sim10^3$ & $\sim10^3$ & $\sim10^3$ & $\sim10^5$ & $\sim10^5$ \\
Number of regimes         & 3 & 3 & 3 & 2 & 4 \\
Metrics             & ADE & ADE & ADE & ADE/FDE & ADE/FDE \\
Dynamic topology   & -- & \checkmark & -- & -- & -- \\
Dynamic edge features & -- & -- & -- & \checkmark & -- \\
Dynamic regime     & -- & -- & \checkmark & -- & \checkmark \\
\bottomrule
\end{tabular}
\caption{\textbf{Summary of dataset properties.} The table highlights complementary dynamics and scales. \checkmark\ indicates presence; -- indicates absence.}
\label{tab:datasummary}
\end{table}
 
\subsubsection{Generated Dataset Construction}
\label{app:generated_datasets}

All generated datasets follow the same high-level protocol. We generate $2000$ graph sequences. For each sequence, the number of nodes is sampled uniformly between $50$ and $100$. Edges are sampled randomly following the network construction procedure used in \citep{networkdynamic}: topology is controlled through the degree structure, and each generated graph is constrained to form a single connected component. This produces random graphs that remain closer to realistic network structure than unconstrained independent edge sampling.

For each node, we create a feature vector of dimension $8$. The node features are initialized independently from a uniform distribution, $x_i^{(0)} \sim \mathcal{U}(0.1,0.6)$. To remain faithful to real world scenarios, we add a Gaussian observational noise $\epsilon_t^i$ at each time step to all node features, with ($\epsilon_t\sim \mathcal{N}(\mu=0,\sigma=0.03)$. For the Linear and Gompertz datasets, each generated graph sequence is assigned a single regime label $r \in \{1,\ldots,K\}$, and the graph is simulated for $10$ time steps according to the corresponding dynamical rule. For the Network dataset, the initial regime is sampled uniformly at the beginning of the simulation. The regime then evolves according to a persistence-based switching process: if the regime has remained unchanged for at least the previous three time steps and that we did not reached yet the horizon, it switches with probability $p=0.3$ to one of the other $K-1$ regimes, sampled uniformly at random; otherwise, the current regime is retained. This construction induces temporally coherent regime segments while allowing stochastic transitions between dynamical modes. No changes of regime occurs during the horizon. We show the dataset generation procedure in the following Algorithm \ref{alg:app_generation}.

\begin{algorithm}[ht]
\caption{Synthetic Graph Sequence Generation}
\label{alg:app_generation}
\begin{algorithmic}[1]
\Require Number of sequences $N_{\mathrm{seq}}=2000$, regimes $K$, switching probability $p=0.3$, context size $C$, horizon $H$
\State $T \leftarrow C+H$
\For{$s=1,\ldots,N_{\mathrm{seq}}$}
    \State Sample $n_s \sim \mathrm{Unif}(\{50,\ldots,100\})$
    \State Sample edges according to the network construction procedure of \citep{networkdynamic}
    \State Initialize $x_i^{(0)} \sim \mathcal{U}(0.1,0.6)$ independently for all nodes $i$

    \If{dataset is Linear or Gompertz}
        \State Sample $r \sim \mathrm{Unif}(\{1,\ldots,K\})$
        \State Set $r_t \leftarrow r$ for all $t=0,\ldots,T-1$
    \ElsIf{dataset is Network}
        \State Sample $r_0 \sim \mathrm{Unif}(\{1,\ldots,K\})$
        \For{$t=1,\ldots,T-1$}
            \If{$t < C$ \textbf{and} $t \geq 3$ \textbf{and} $r_{t-1}=r_{t-2}=r_{t-3}$}
                \State With probability $p$, sample
                $
                r_t \sim \mathrm{Unif}\bigl(\{1,\ldots,K\}\setminus\{r_{t-1}\}\bigr)
                $
            \Else
                \State Set $r_t \leftarrow r_{t-1}$
            \EndIf
        \EndFor
    \EndIf

    \For{$t=0,\ldots,T-1$}
        \State Simulate the graph dynamics at time $t$ under regime $r_t$
    \EndFor
\EndFor
\end{algorithmic}
\end{algorithm}

\paragraph{Linear Dynamics}
\label{app:linear}

The linear dataset removes interaction effects so that regime recovery depends on aggregating temporal evidence across independent entities. Each node follows
\begin{equation}
x_{t+1}^i = a_k x_t^i+b_k,
\qquad
k\in\{1,\ldots,K\},
\end{equation}
where $(a_k,b_k)$ are regime-specific coefficients. All nodes in a sequence share the same regime. This dataset serves as a sanity check: a model should recover the shared dynamical mode without relying on relational information. Note that the linearity is in the dynamics, not in the raw features evolution. If we unravel the equation and neglect the bias term, the features evolve according to an exponential law: $x_t^i=x_0(a_k)^t$.

\begin{table}[ht]
\captionsetup{skip=10pt}
\centering
\small
\begin{tabular}{llll}
\toprule
Regime type & $a_k$ & $b_k$ & Behavior\\
\midrule
Decay & 0.8 & 0 & Exponential decay \\
Stable & 0.0 & 1 & Constant drift independent of features value \\
Increase & 1.2 & 0 & Exponential increase \\
\bottomrule
\end{tabular}
\caption{\textbf{Linear dataset regimes.} The three dynamics corresponds to different linear evolution patterns.}
\label{tab:linear}
\end{table}

\paragraph{Gompertz Dynamics with Topological Events}
\label{app:gompertz}

The Gompertz dataset introduces non-linear growth and local topological changes. Node features represent scalar volumes and evolve according to a discrete Gompertz-type update:
\begin{equation}
x_{t+1}^i
=
a_k^{1-\exp(-c_k)}
\left(x_t^i\right)^{\exp(-c_k)},
\qquad
k\in\{1,\ldots,K\}.
\end{equation}
The regime controls the growth parameters. When a node reaches the saturation threshold $\theta_{\mathrm{thr}}=0.5$, it is removed and replaced by two newly spawned entities. The two entities connectivity is the same as the one of its parent. This induces a local change in the node set and graph topology. The setting tests whether regime inference remains stable under non-linear feature evolution and changing graph support. We list the set of considered regimes and their associated parameters in Table \ref{tab:app_gompertz}.

\begin{table}[ht]
\captionsetup{skip=10pt}
\centering
\small
\begin{tabular}{llll}
\toprule
Regime type & $a_k$ & $c_k$ & Behavior\\
\midrule
Tumor regression & 0.05 & 0.08 & Shrinks toward small volume \\
Stable disease & 0.5 & 0.00 & Approximately constant \\
Progressive disease & 1 & 0.08 & Grows beyond initial features range \\
\bottomrule
\end{tabular}
\caption{\textbf{Gompertz regimes.} The three dynamics corresponds to biological relevant tumor growth patterns, namely regression, stable and progression.}
\label{tab:app_gompertz}
\end{table}

\paragraph{Network Interaction Dynamics}
\label{app:network}

The network dataset is designed to test whether the model can identify regimes defined by interaction mechanisms rather than by isolated node trajectories. The same graph topology can produce different trajectories depending on how node states propagate over edges. We use three dynamical families from \citep{networkdynamic}, corresponding to degree-averting, degree-driven, and homogeneous information flow.

For a weighted adjacency matrix $A$, the node state $x_i(t)$
evolves according to one of three continuous-time systems.

The regulatory dynamics ($\mathbb{R}_2$) combine linear decay with a
saturating input from neighboring nodes:
\begin{equation}
    \frac{dx_i(t)}{dt}
    =
    -x_i(t)
    +
    \sum_{j=1}^{N}
    A_{ij}
    \frac{x_j(t)^2}{1+x_j(t)^2}.
\end{equation}
The saturating input limits the influence of large neighboring
states and corresponds to degree-averting flow, in which
information tends to avoid hubs and favor peripheral pathways.

The population dynamics ($\mathbb{P}$) combine self-limiting decay
with positive reinforcement from neighboring nodes:
\begin{equation}
    \frac{dx_i(t)}{dt}
    =
    -x_i(t)^3
    +
    \sum_{j=1}^{N}
    A_{ij}x_j(t)^2.
\end{equation}
This positive reinforcement yields degree-driven flow,
in which hubs dominate propagation.

The epidemic dynamics ($\mathbb{E}$) combine recovery with infection
pressure from neighboring nodes, modulated by the susceptible
fraction $(1-x_i(t))$:
\begin{equation}
    \frac{dx_i(t)}{dt}
    =
    -x_i(t)
    +
    \sum_{j=1}^{N}
    A_{ij}(1-x_i(t))x_j(t).
\end{equation}
The susceptibility factor modulates the influence of infected
neighbors, yielding homogeneous flow in which nodes contribute
more uniformly across the graph.

We discretize the continuous flow equations by integrating
the dynamics with a $\Delta t = 0.01$. A summary of each dynamics is provided in Table \ref{tab:network_dynamics}.
 
\begin{table}[ht]
\captionsetup{skip=10pt}
\centering
\small
\setlength{\tabcolsep}{3pt}
\resizebox{\columnwidth}{!}{%
\begin{tabular}{llll}
\toprule
Family & Flow class & Regime signal & Propagation pattern \\
\midrule
$\mathbb{R}_2$ & Saturating regulatory influence & Degree-averting & Peripheral pathways dominate \\
$\mathbb{P}$ & Self-limited growth with neighbor reinforcement & Degree-driven & Hubs dominate propagation \\
$\mathbb{E}$ & Recovery and susceptibility-limited contagion & Homogeneous & Propagation is broadly distributed \\
\bottomrule
\end{tabular}}
\caption{\textbf{Network interaction regimes.} The three network dynamics differ in the mechanism by which information propagates over the same graph topology.}
\label{tab:network_dynamics}
\end{table}

This dataset is the most direct synthetic test of graph-aware regime inference. A velocity-only method may miss these regimes because the distinction lies in how the graph mediates propagation, not only in marginal feature changes.

\subsubsection{Real-World Datasets}
\label{app:real_world}

\paragraph{SportVU NBA}
\label{app:nba}

We use the SportVU NBA player-tracking dataset to evaluate regime-aware forecasting in a multi-agent setting with strong interaction structure. The dataset consists of professional basketball tracking sequences annotated with play outcomes. We consider two outcome categories: scoring plays, which end in a made basket, and rebounding plays, which follow a missed shot. These outcome categories define sequence-level regimes, which remain fixed over the prediction horizon.

The dataset is available in \citep{nba1}. We use the train/test split provided by the authors of \citep{nba2}. To construct the final benchmark, we pool scoring and rebounding sequences while preserving the original train/test partition. No additional outcome relabeling is performed. The interaction graph is defined as a fully connected graph. For each node, we use the spatial coordinates $(x,y)$ together with a static entity-type descriptor indicating whether the node corresponds to a player or the ball.  Edge features encode pairwise relational attributes, including the $\mathcal{L}_2$ distance between entities and a team-membership indicator: $1$ for entities on the same team, $0$ for player--ball edges, and $-1$ for players on opposing teams. This indicator is treated as a $\mathbb{Z}_2$-symmetric relational field. The edge $\mathcal{L}_2$ distance and spatial coordinates are the only time-varying features. The preprocessing pipeline used to construct the trajectories and regime labels is provided in our code.

\paragraph{AMASS with BABEL Annotations}
\label{app:amass}

We use AMASS with BABEL annotations to evaluate regime-aware forecasting under time-varying action structure. AMASS provides articulated human motion-capture trajectories, while BABEL provides temporally aligned action annotations. In contrast to SportVU NBA, where the regime is fixed for an entire sequence, BABEL action labels may change within a sequence. This setting therefore evaluates whether a model can track changes in the latent action regime over time while forecasting joint-level motion.

The AMASS motion data are available from \citep{amass_homepage}. Since AMASS aggregates multiple motion-capture sources, we restrict our experiments to four large and commonly used subsets: BMLmovi and BMLrub \citep{bml}, CMU \citep{cmu}, and KIT \citep{kit}. We pair these motion sequences with BABEL annotations, available from \citep{babel_webportal}, and map fine-grained action labels to a smaller set of dominant coarse regimes. Specifically, as shown in Table \ref{tab:app_babel} we retain four action groups: step/walk, stand, turn, and sit/lower. We drop all the time steps or motion sequences that had no actions falling in the associated pooling. The stemming, grouping, filtering, and trajectory preprocessing steps are provided in our code. Unless otherwise stated, action labels are used only for evaluation. The graphs are created using the SMPL 22 joints skeleton \citep{smpl}. For each node, we use the spatial coordinates $(x,y)$ together with static anatomical descriptors: joint type, left/right laterality, and distance to the pelvis. Edge features represent static relational attributes, including limb membership (e.g., arm or leg), fixed inter-joint distance, and left/right laterality. The spatial coordinates are the only time-varying features.

\begin{table}[ht]
\captionsetup{skip=10pt}
\centering
\small
\begin{tabular}{llll}
\toprule
Regime type & Example of Action Names & Global Motion Pattern \\
\midrule
Step/Walk & `walk away`, `slow walk`, `step forward`, ... & Directional feet movement \\
Stand & `look`, `poses`, `wait`, ...  & Most body static \\
Turn & `twist body to the left`, `rotate`, `twirl`, ... & Body rotating \\
Sit/Lower & `sit`, `lean`, `crouch`, ...  & Body joints lowering to the ground \\
\bottomrule
\end{tabular}
\caption{\textbf{AMASS/BABEL regimes.} The four dynamics corresponds to dominant motion patterns in the AMASS \citep{amass} dataset aligned with BABEL labels \citep{babel}.}
\label{tab:app_babel}
\end{table}

\subsection{Complementary Experiments}

\subsubsection{Training results and statistical significance} \label{app:std}

Training reported as the results of the four folds cross validation with best hyperparameters.

\begin{table*}[ht]
\captionsetup{skip=10pt}
\centering
\setlength{\tabcolsep}{3pt}
\resizebox{\textwidth}{!}{%
\begin{tabular}{lccccc}
\toprule
 & \multicolumn{3}{c}{Synthetic} & \multicolumn{2}{c}{Real-world} \\
\cmidrule(lr){2-4} \cmidrule(lr){5-6}
Model 
& Linear & Gompertz \citep{gompertz1} & Network \citep{networkdynamic} & NBA \citep{nba1} & AMASS \citep{amass} \\
\midrule
Persistence (Static)   
& 5.518 ($\pm$0.055) 
& 2.498 ($\pm$0.025) 
& 0.103 ($\pm$0.001) 
& 4.51 ($\pm$0.07) / 7.78 ($\pm$0.13) 
& 0.092 ($\pm$0.002) / 0.133 ($\pm$0.003) \\
Persistence (Velocity) 
& 2.911 ($\pm$0.029) 
& 2.881 ($\pm$0.029) 
& 0.057 ($\pm$0.001) 
& 2.51 ($\pm$0.05) / 5.81 ($\pm$0.10) 
& 0.131 ($\pm$0.003) / 0.276 ($\pm$0.006) \\
\midrule
S4 \citet{s4} 
& 4.156 ($\pm$0.125) 
& 4.378 ($\pm$0.131) 
& 0.165 ($\pm$0.006) 
& 2.13 ($\pm$0.08) / 5.04 ($\pm$0.18) 
& 0.428 ($\pm$0.018) / 0.624 ($\pm$0.028) \\
rSLDS \citet{rslds} 
& 2.720 ($\pm$ 0.058)
& 3.133 ($\pm$ 0.047)
& 0.054 ($\pm$ 0.001) 
& 2.93 ($\pm$ 0.02) / 5.40 ($\pm$ 0.05)
& 0.073 ($\pm$ 0.004) / 0.133 ($\pm$ 0.006) \\
RED-SDS \citet{redsds} 
& 3.388 ($\pm$ 0.031)
& 3.152 ($\pm$ 0.128)
& 0.046 ($\pm$ 0.002) 
& 4.72 ($\pm$ 0.07)  / 8.36 ($\pm$ 0.13)
& 0.071 ($\pm$ 0.002) / 0.119 ($\pm$ 0.002) \\
TESTAM \citet{testam} 
& 2.923 ($\pm$ 0.198) 
& 2.679 ($\pm$ 0.165) 
& 0.042 ($\pm$ 0.002) 
& 2.55 ($\pm$ 0.09) / 4.57  ($\pm$ 0.21)
& 0.051 ($\pm$ 0.004) / 0.093 ($\pm$ 0.003)\\
EvolveGCN \citep{ref_evolve_gcn} 
& 3.398 ($\pm$0.071) 
& 2.654 ($\pm$0.054) 
& 0.056 ($\pm$0.001) 
& 1.74 ($\pm$0.06) / 4.97 ($\pm$0.15) 
& 0.059 ($\pm$0.002) / 0.087 ($\pm$0.003) \\
LaM-SLidE \citet{lam-slide} 
& \underline{2.083} ($\pm$0.043) 
& N/A 
& 0.053 ($\pm$0.001) 
& \textbf{0.74} ($\pm$0.03) / \textbf{1.30} ($\pm$0.06) 
& 0.062 ($\pm$0.004) / 0.085 ($\pm$0.003) \\
\midrule
GraphSSM \citet{graphssm} 
& 2.521 ($\pm$0.052) 
& 2.156 ($\pm$0.046) 
& 0.050 ($\pm$0.001) 
& 1.66 ($\pm$0.06) / 4.62 ($\pm$0.15) 
& 0.057 ($\pm$0.002) / 0.083 ($\pm$0.003) \\
GraphSSM+HERON \citet{graphssm} 
& 2.481 ($\pm$0.049) 
& \underline{2.061} ($\pm$0.041) 
& \underline{0.046} ($\pm$0.001) 
& 1.58 ($\pm$0.05) / 4.39 ($\pm$0.13) 
& \underline{0.054} ($\pm$0.002) / \underline{0.079} ($\pm$0.003) \\
\midrule
HERON 
& \textbf{1.921} ($\pm$0.038) 
& \textbf{1.945} ($\pm$0.038) 
& \textbf{0.038} ($\pm$0.001) 
& \underline{0.80} ($\pm$0.03) / \underline{1.51} ($\pm$0.06) 
& \textbf{0.046} ($\pm$0.003) / \textbf{0.067} ($\pm$0.002) \\
\bottomrule
\end{tabular}}

\caption{\textbf{Regime-conditioned forecasting performance.} Train ADE is reported for generated datasets, and ADE/FDE for motion datasets. Values are mean ($\pm$ standard deviation) over 4 folds. Best results are in bold and second-best underlined. N/A means that the method was not applicable.}
\label{tab:train_forecasts}
\end{table*}

\begin{table}[ht]
\captionsetup{skip=10pt}
\centering
\setlength{\tabcolsep}{4pt}

\resizebox{\columnwidth}{!}{%
\begin{tabular}{lccccc}
\toprule
 & \multicolumn{3}{c}{Synthetic} & \multicolumn{2}{c}{Real-world} \\
\cmidrule(lr){2-4} \cmidrule(lr){5-6}
Method & Linear & Gompertz & Network & NBA & AMASS \\
\midrule
Latent-HMM 
& \textbf{0.979} ($\pm$0.051) 
& \textbf{0.965} ($\pm$0.044) 
& \underline{0.716} ($\pm$0.023) 
& \underline{0.604} ($\pm$0.016) 
& 0.329 ($\pm$0.017) \\
rSLDS \citet{rslds}
& \underline{0.958} ($\pm$0.059) 
& \underline{0.946} ($\pm$0.016) 
& 0.331 ($\pm$0.024) 
& 0.362 ($\pm$0.019) 
& \underline{0.330} ($\pm$0.025) \\
RED-SDS \citet{redsds}
& 0.946 ($\pm$0.026) 
& 0.939 ($\pm$0.019) 
& 0.570 ($\pm$0.007) 
& 0.493 ($\pm$0.023) 
& 0.318 ($\pm$0.017) \\
\midrule
HERON 
& 0.948 ($\pm$0.031) 
& 0.931 ($\pm$0.012) 
& \textbf{0.920} ($\pm$0.009) 
& \textbf{0.878} ($\pm$0.011) 
& \textbf{0.419} ($\pm$0.018) \\
\bottomrule
\end{tabular}%
}

\caption{\textbf{Self-supervised regime recovery.} Train macro-F1 quantifies how well the learned regimes align with ground-truth dynamical labels, which are used only for evaluation.
Best results are in bold and second-best results are underlined. Values are mean ($\pm$ standard deviation) over 4 folds.}
\label{tab:regime-metric-std}
\end{table}

\subsubsection{Quantitative regime interventions}\label{app:interventions}

We test whether HERON's learned regimes encode functionally distinct dynamics by intervening on the regime used during rollout. For each ground-truth regime $(j)$, we roll out the corresponding sequences while clamping the model to regime $(i)$, average the resulting ADE, and normalize each column by the matched intervention $(i=j)$ (diagonal $=1$; lower is better). The matched regime achieves the lowest ADE on every dataset. Incorrect interventions increase error by 15--83\% on Network and 16--23\% on NBA, indicating that the experts encode non-interchangeable relational dynamics rather than redundant parameterizations. The effect is smaller on AMASS (2--31\%), consistent with its smooth and partially overlapping motion dynamics. Nevertheless, clearly distinct modes remain well separated: for example, intervening between Stand and Lower increases ADE by 28--31\%. See Table \ref{tab:rebuttal_interventions} for the extensive list of all results. Overall, changing the inferred regime systematically alters the predicted dynamics, while the matched regime yields the best single-regime rollout, providing quantitative evidence that HERON's regimes meaningfully govern the generated trajectories.

\begin{table}[ht]
\centering
\small
\setlength{\tabcolsep}{5pt}

\begin{tabular}{lccc}
\toprule
\multicolumn{4}{l}{\textbf{Network}} \\
\cmidrule(lr){1-4}
& \multicolumn{3}{c}{\textit{True regime}} \\
\cmidrule(lr){2-4}
\textit{Clamped regime} & Regulatory & Epidemic & Mutualistic \\
\midrule
Regulatory  & 1.00 & 1.42 & 1.19 \\
Epidemic    & 1.83 & 1.00 & 1.29 \\
Mutualistic & 1.21 & 1.15 & 1.00 \\
\bottomrule
\end{tabular}

\vspace{0.6em}

\begin{tabular}{lcccc}
\toprule
\multicolumn{5}{l}{\textbf{AMASS}} \\
\cmidrule(lr){1-5}
& \multicolumn{4}{c}{\textit{True regime}} \\
\cmidrule(lr){2-5}
\textit{Clamped regime} & Rotation & Step & Stand & Lower \\
\midrule
Rotation & 1.00 & 1.03 & 1.09 & 1.04 \\
Step     & 1.02 & 1.00 & 1.15 & 1.14 \\
Stand    & 1.07 & 1.19 & 1.00 & 1.31 \\
Lower    & 1.17 & 1.11 & 1.28 & 1.00 \\
\bottomrule
\end{tabular}

\vspace{0.6em}

\begin{tabular}{lcc}
\toprule
\multicolumn{3}{l}{\textbf{NBA}} \\
\cmidrule(lr){1-3}
& \multicolumn{2}{c}{\textit{True regime}} \\
\cmidrule(lr){2-3}
\textit{Clamped regime} & Basket & Rebound \\
\midrule
Basket  & 1.00 & 1.16 \\
Rebound & 1.23 & 1.00 \\
\bottomrule
\end{tabular}
\vspace{7pt}
\caption{\textbf{Forecasting results with regime clamping.}
Rows indicate the regime imposed during forecasting, while columns indicate the
true regime. ADE is normalized column-wise by the matched intervention, such
that diagonal entries equal 1.}
\label{tab:rebuttal_interventions}
\end{table}

We further evaluate whether regime interventions preserve physically plausible rollouts, i.e., whether alternative regimes modify the dynamics without producing anatomically inconsistent human motion on AMASS or implausible player trajectories on NBA (see Table \ref{tab:rebuttal_intervention_amass_constraints} and \ref{tab:rebuttal_intervention_nba_constraints} respectively). On AMASS, we measure relative bone-length violation, which captures deviations from skeletal consistency, and per-joint jerk normalized by real motion, which measures whether predicted trajectories remain comparably smooth. On NBA, we report the fraction of predicted positions outside the court and player acceleration relative to real trajectories, capturing spatial validity and kinematic realism. All reported statistics are computed over the test set. Although incorrect regime interventions increase ADE, their constraint statistics remain close to those of matched-regime rollouts: bone-length violation changes from 2.51\% to 2.66\% and normalized jerk from 2.89 to 2.87 on AMASS, while the out-of-court rate changes from 3.37\% to 3.65\% and normalized acceleration from 1.19 to 1.20 on NBA. Thus, clamping an alternative regime changes the predicted dynamics without substantially reducing their physical or kinematic plausibility. Combined with the ADE intervention results, this shows that HERON produces distinct yet similarly constraint-compliant dynamical alternatives. These are interventions on the model's latent regime, not identified causal interventions on the real-world process; they quantify whether the learned experts control distinct predictive dynamics.

\begin{table}[ht]
\centering
\small
\begin{tabular}{lcc}
\toprule
AMASS condition & Bone-length violation (\%) & Motion jerk ($\times$ real) \\
\midrule
Ground-truth future & 0.00 & 1.00 \\
Clamp = correct regime & 2.51 & 2.89 \\
Clamp = wrong regime & 2.66 & 2.87 \\
\bottomrule
\end{tabular}
\vspace{7pt}
\caption{\textbf{Constraint violation with regime clamping on AMASS.}}
\label{tab:rebuttal_intervention_amass_constraints}
\end{table}

\begin{table}[ht]
\centering
\small
\begin{tabular}{lcc}
\toprule
NBA condition & Out-of-court (\%) & Player acceleration ($\times$ real) \\
\midrule
Ground-truth future & 2.21 & 1.00 \\
Clamp = correct regime & 3.37 & 1.19 \\
Clamp = wrong regime & 3.65 & 1.20 \\
\bottomrule
\end{tabular}
\vspace{7pt}
\caption{\textbf{Constraint violation with regime clamping on NBA.}}
\label{tab:rebuttal_intervention_nba_constraints}
\end{table}

\subsubsection{Imbalanced regime frequencies}\label{app:imbalance}

We evaluate HERON on a controlled variant of the Network dataset with regime frequencies 70\%/20\%/10\%. Under the original balance loss, which matches aggregate regime usage to a uniform distribution, HERON fails to recover the rarest regime (minority recall 0.00). We therefore replace the fixed uniform target with a categorical prior learned jointly from the training data, without regime supervision. This modification substantially recovers the minority regime (recall 0.70) and improves macro-F1 from 0.60 to 0.78 and regime AUC from 0.62 to 0.97, while also reducing ADE from 0.049 to 0.044 (see Table \ref{tab:rebuttal_imbalance}). These results show that replacing the fixed uniform prior with a learned one substantially improves HERON under strongly imbalanced regimes, suggesting that the uniform regime-distribution assumption can be relaxed.

\begin{table}[ht]
\centering
\small
\begin{tabular}{lcccc}
\toprule
Model & Test macro-F1 & Minority recall & Regime AUC & ADE \\
\midrule
Default HERON (uniform balance) & 0.60 & 0.00 & 0.62 & 0.049 \\
HERON + learnable prior & 0.78 & 0.70 & 0.97 & 0.044 \\
\bottomrule
\end{tabular}
\vspace{7pt}
\caption{\textbf{Imbalanced regimes.} Controlled Network variant with regime frequencies 70\%/20\%/10\%.}
\label{tab:rebuttal_imbalance}
\end{table}

\subsubsection{Overlapping regimes and soft routing}\label{app:multik}

To test whether the one-active-regime assumption is restrictive, we generate sequences from convex mixtures of two regimes with known weights $w=(\lambda,1-\lambda)$, and compare the original hard rollout gate, $r=\mathrm{onehot}(\arg\max(p))$, with soft routing, $r=p$. The hard- and soft-routing variants are trained separately under the same protocol (see Table \ref{tab:rebuttal_overlap} for detailed results). Both recover the mixture similarly (posterior--mixture agreement accuracy: 0.85 vs. 0.87), but soft routing better preserves it during generation: rollout-weight error decreases from 0.28 to 0.13 ($-54\%$), and ADE on mixed sequences improves from 2.02 to 1.85 ($-8.4\%$). The trade-off on pure sequences is modest (ADE 2.22 to 2.32, $+4.5\%$), where hard selection correctly matches a single active regime. Thus, HERON's posterior already captures mixed regime membership. The limitation arises from collapsing it to one expert at rollout. Posterior-weighted routing, which combines expert predictions according to the inferred regime probabilities, provides a straightforward extension for settings with coexisting dynamics. This controlled experiment addresses simultaneous mixtures; continuous temporal transitions remain approximated under the current discrete prior, although soft routing offers one practical extension.

\begin{table}[ht]
\centering
\small
\begin{tabular}{lcccc}
\toprule
Setting & ADE (pure) & ADE (mixed) & Posterior calibration & Rollout-weight error \\
\midrule
HERON -- hard gate & 2.22 & 2.02 & 0.85 & 0.28 \\
HERON -- soft gate & 2.32 & 1.85 & 0.87 & 0.13 \\
\bottomrule
\end{tabular}
\vspace{7pt}
\caption{\textbf{Overlapping regimes.} Hard versus posterior-weighted soft routing.}
\label{tab:rebuttal_overlap}
\end{table}

\subsubsection{AMASS regime-recovery errors}\label{app:amass_errors}

AMASS recovery is primarily limited near action boundaries, where BABEL's semantic labels are an imperfect proxy for instantaneous motion dynamics. In particular, the labeling frequently uses the Transition label for short and heterogeneous changes between most action pairs and therefore does not define a coherent dynamical mode. In the preprocessing pipeline used for the main-paper results, we exclude this label from the retained action set and map its frames to the preceding action. Moreover, physical motions change smoothly, whereas annotations switch discretely, so HERON generally requires several observations before identifying the new mode. Consistent with this explanation, 38\% of errors occur within 5 frames of a label transition, and macro-F1 increases from 0.412 overall to 0.566 on stable motion segments. Manual inspection also reveals semantic boundary ambiguities, such as turning motions being grouped with either rotation or stepping. These results indicate that label-dynamics mismatch is an important contributor to the lower aggregate recovery score, particularly near ambiguous transitions, while the remaining errors on stable segments also reflect residual model limitations.

\subsubsection{Regime reproducibility across runs}\label{app:regime_stability}

To test whether HERON recovers reproducible dynamical modes rather than seed-specific partitions, we trained five independent seeds and evaluated all ten run pairs on the same held-out sequences. We report mean $\pm$ standard deviation across these ten pairwise comparisons for: (i) ARI, which measures agreement between frame partitions independently of regime labels and adjusts for chance; (ii) matched frame accuracy, the fraction of frames assigned to the same regime after a single global Hungarian alignment; and (iii) worst-mode Jaccard, the minimum intersection-over-union between the frame sets assigned to any pair of matched regimes, thereby exposing instability in the least reproducible mode. Null ARI is the 95th percentile over 1,000 circular-shift surrogates, which destroy cross-run alignment while preserving regime frequencies and segment durations. Because pairs sharing a run are not statistically independent, these statistics characterize cross-seed variability rather than independent-sample uncertainty. Results are presented in Table \ref{tab:rebuttal_regime_stability}.

\begin{table}[ht]
\centering
\small
\begin{tabular}{lccccc}
\toprule
Dataset & $K$ & Inter-run ARI & Matched accuracy & Worst-mode Jaccard & Null ARI (95th \%ile) \\
\midrule
Network & 3 & $0.92\pm0.02$ & $0.98\pm0.02$ & $0.96\pm0.03$ & 0.005 \\
NBA & 2 & $0.83\pm0.04$ & $0.96\pm0.01$ & $0.90\pm0.02$ & 0.004 \\
AMASS & 4 & $0.48\pm0.07$ & $0.61\pm0.08$ & $0.57\pm0.05$ & 0.005 \\
\bottomrule
\end{tabular}
\vspace{7pt}
\caption{\textbf{Cross-seed regime reproducibility.} Statistics over all ten pairs from five independent seeds.}
\label{tab:rebuttal_regime_stability}
\end{table}

HERON is highly consistent on Network and NBA, reaching ARI 0.92/0.83, matched accuracy 0.98/0.96, and worst-mode Jaccard 0.96/0.90, respectively. Consistency is lower on AMASS, in line with its weaker alignment to the coarse semantic labels and the ambiguity induced by smooth motion transitions. Nevertheless, independent runs recover meaningfully similar partitions: ARI reaches 0.48, compared with a temporal-null threshold of 0.005, while matched accuracy and worst-mode Jaccard remain 0.61 and 0.57. Thus, HERON repeatedly discovers similar dynamical patterns on AMASS, although this structure is only partially aligned with the four annotated action classes.

\subsubsection{GraphSSM enhanced with HERON}
\label{app:graphssm-HERON}

\paragraph{Motivation.} Section~\ref{sec:architecture} claims that the HERON regime pathway can wrap any latent graph-temporal backbone exposing an entity-level state. We instantiate this on GraphSSM~\citep{graphssm}, where each node trajectory is propagated by a structured SSM kernel with graph-based mixing. GraphSSM is competitive on local trajectories (Table~\ref{tab:train_forecasts}) but offers no system-level mode: in regime-structured data, the active mode is silently absorbed into the per-node SSM state, foreclosing regime tracking and the interventional analysis of Section~\ref{sec:result_regime}.

\paragraph{Component mapping.} We identify the GraphSSM hidden state with the entity-level deterministic state $h_t^i$ of Section~\ref{sec:state_representation}, and let the GraphSSM update play the role of the entity-level transition. Three HERON modules are added around it:
\begin{enumerate}
    \item A graph-level GRU $h_t^G \in \mathbb{R}^{d_H}$ and a categorical regime $r_t \in \{1,\dots,K\}$.
    \item Hierarchical regime inference: attention pooling~\citep{attpooling} over the GraphSSM node embeddings produces $\mathrm{emb}_t^G$; the posterior $q_\phi(r_t \mid h_t^G, \mathrm{emb}_t^G)$ is sampled via Gumbel--Softmax with straight-through estimator; a separate $\psi$-module supplies the prior $p_\psi(\hat r_t \mid h_t^G, \widehat{\mathrm{emb}}_t^G)$ used at rollout.
    \item A regime-gated MoE producing the entity-level latent $z_t^i$, with one expert per regime.
\end{enumerate}
The forward pass is Algorithm~\ref{alg:HERON} with the GraphSSM kernel substituted for the entity-level transition operator.

\paragraph{Implementation.} We use GraphSSM-S4~\citep{graphssm} stacked to the depth of our HERON encoder. Regime cardinality matches standalone HERON ($K{=}3$ synthetic, $K{=}2$ NBA, $K{=}4$ AMASS). All non-backbone hyperparameters follow Table~\ref{tab:hyper}. We explicitely matched the capacity of GraphSSM and its counterpart with the HERON module on top, so that any gain in performances cannot be attributed to a pure increase in capacity.

\paragraph{Forecasting.} GraphSSM+HERON improves over vanilla GraphSSM on every benchmark (Table~\ref{tab:train_forecasts}: ADE drops by $1.7\%$ (Linear), $6.2\%$ (Gompertz), $4.1\%$ (Network); ADE/FDE drop by $3.6\%/3.5\%$ (NBA) and $1.8\%/2.4\%$ (AMASS). The largest gains arise on the most clearly regime-structured benchmarks (Gompertz, Network), supporting the central claim that an explicit regime variable pays off when latent dynamical modes are present. GraphSSM+HERON does \emph{not} match standalone HERON (e.g.\ $0.047$ vs.\ $0.038$ on Network), indicating that the SSM kernels and the regime-conditioned MoE supply complementary rather than redundant inductive biases.

\paragraph{Regime tracking.}

Table \ref{tab:graphssm_regime} shows that plugging HERON on top of GraphSSM allowed to retrieve regimes, even if performances are slightly degraded compared to pure HERON.

\begin{table}[ht]
\captionsetup{skip=10pt}
\centering
\small
\begin{tabular}{lccccc}
\toprule
 & \multicolumn{3}{c}{Synthetic} & \multicolumn{2}{c}{Real-world} \\
\cmidrule(lr){2-4} \cmidrule(lr){5-6}
Metric & Linear & Gompertz & Network & NBA & AMASS \\
\midrule
Macro-F1 & 0.766 & 0.703 & 0.790 & 0.872 & 0.32 \\
AUC      & 0.703 & 0.944 & 0.868 & 0.899 & 0.408 \\
\bottomrule
\end{tabular}
\caption{\textbf{Self-supervised regime recovery.} Test macro-F1 and AUC quantify how well learned regimes align with ground-truth dynamical labels, which are used only for evaluation.}
\label{tab:graphssm_regime}
\end{table}

\paragraph{Takeaway.} The regime interface adds as a lightweight wrapper to a strong latent graph-temporal backbone without rewriting it, yielding concurrent gains in forecasting and regime tracking.

\subsubsection{Sensitivity to mis-specification of K}\label{app:k_sensitivity}

To assess HERON's sensitivity to misspecification of the number of regimes, we retrained the model with $K$ below and above the reference value. Forecasting is substantially less sensitive to $K$ than regime recovery (see Table \ref{tab:rebuttal_k_sensitivity} for detailed results). On Network, over-specifying $K=4$ preserves the reference-$K$ ADE of 0.038, while only three experts remain active. Even at $K=5$, ADE is 0.041, with only three experts used. Similarly, NBA models with $K=3$ or $4$ activate only two experts, obtaining ADE 0.89 and 0.84, respectively, versus 0.81 at $K=2$. Under-specification affects regime identification more strongly. On Network, $K=2$ increases ADE moderately from 0.038 to 0.043, but reduces macro-F1 from 0.91 to 0.55. On AMASS, $K=3$ gives the best ADE/FDE, whereas the reference $K=4$ gives the best macro-F1/AUC. This likely reflects AMASS's smooth transitions and partially overlapping motion patterns, for which the number of predictive dynamical modes need not coincide with the number of semantic action labels. Overall, forecasting is relatively robust to moderate over-specification, whereas reference-label-aligned recovery remains sensitive to both under- and over-specification. We therefore use active-expert utilization together with the validation criterion described in Appendix~\ref{app:k} to select $K$ in practice.

\begin{table*}[ht]
\centering
\small
\setlength{\tabcolsep}{3pt}

\resizebox{\textwidth}{!}{%
\begin{tabular}{lccccc@{\hspace{1cm}}lcccccc}
\toprule
\multicolumn{6}{c}{Network} & \multicolumn{7}{c}{NBA} \\
$K$ & ADE & macro-F1 & AUC & Experts used & & $K$ & ADE & FDE & macro-F1 & AUC & Experts used & \\
\midrule
2 & 0.043 & 0.55 & 0.75 & 2 & & \textbf{2} & \textbf{0.81} & \textbf{1.49} & \textbf{0.89} & \textbf{0.91} & 2 & \\
\textbf{3} & \textbf{0.038} & \textbf{0.91} & \textbf{0.97} & 3 & & 3 & 0.89 & 1.73 & 0.83 & 0.86 & 2 & \\
4 & 0.038 & 0.81 & 0.88 & 3 & & 4 & 0.84 & 1.67 & 0.87 & 0.87 & 2 & \\
5 & 0.041 & 0.73 & 0.76 & 3 & & & & & & & & \\
\bottomrule
\end{tabular}}

\vspace{0.4em}

\resizebox{0.5\textwidth}{!}{%
\begin{tabular}{lccccc}
\toprule
\multicolumn{6}{c}{AMASS} \\
$K$ & ADE & FDE & macro-F1 & AUC & Experts used \\
\midrule
2 & 0.051 & 0.080 & 0.27 & 0.49 & 2 \\
3 & \textbf{0.044} & \textbf{0.059} & 0.29 & 0.48 & 3 \\
\textbf{4} & 0.047 & 0.066 & \textbf{0.41} & \textbf{0.57} & 4 \\
5 & 0.051 & 0.085 & 0.27 & 0.52 & 5 \\
\bottomrule
\end{tabular}}

\caption{\textbf{Sensitivity to misspecification of the number of regimes.} Bold $K$ marks the reference regime count; metric bolding follows the best value in each dataset.}
\label{tab:rebuttal_k_sensitivity}
\end{table*}

\subsubsection{Selecting the number of regimes without labels}\label{app:k}

To select $K$ without regime labels or domain knowledge, we train HERON for $K\in\{1,\ldots,6\}$ and evaluate the predictive NLL of held-out validation futures. We then select the elbow using a fixed rule: after rescaling the curve, we choose the point of maximum deviation from the line joining its endpoints, marking where further increases in $K$ yield diminishing returns. This recovers the reference $K=3/2/4$ on Network/NBA/AMASS, whereas the paper's velocity heuristic selects 2/4/5 (results presented in Table \ref{tab:rebuttal_k_selection}).

As a complementary diagnostic, we estimate the effective $K$ of an overcomplete HERON model trained with $K=6$. We first remove experts whose usage falls below 10\%, then compare the remaining experts through rollouts clamped from the same validation states. We normalize pairwise forecast disagreements by the median among surviving experts, so 1 denotes a typical pair. Under the primary rule, pairs below 0.5 are connected in a redundancy graph, and connected components define the effective $K$. This yields 3/2/5 distinct experts. On AMASS, the additional effective mode suggests specialization to locally distinct motion patterns, while held-out likelihood still favors $K=4$. Thus, predictive likelihood provides a practical label-free selection rule, while effective $K$ identifies inactive or dynamically redundant experts.

\begin{table}[ht]
\centering
\small
\begin{tabular}{lccc}
\toprule
Selection rule & Network ($K=3$) & NBA ($K=2$) & AMASS ($K=4$) \\
\midrule
Expert usage & 3 & 2 & 6 \\
Effective-$K$ & 3 & 2 & 5 \\
Held-out likelihood & \textbf{3} & \textbf{2} & \textbf{4} \\
Velocity baseline & 2 & 4 & 5 \\
\bottomrule
\end{tabular}
\vspace{7pt}
\caption{\textbf{Label-free $K$ selection and diagnostics.} Reference regime counts are shown in parentheses.}
\label{tab:rebuttal_k_selection}
\end{table}

\subsection{Hyperparameters}
\label{app:hyperparameters}

We list in this section the set of optimal hyperparameters for the framework that resulted from the random search performed with the cross-validation. 

\begin{table}[ht]
\captionsetup{skip=10pt}
\centering
\small
\begin{tabular}{llll}
\toprule
\textbf{Hyperparameter} & \textbf{Synthetic Datasets} & \textbf{NBA Dataset } & \textbf{AMASS Dataset }\\
\midrule
batch size & 256 & 64 & 128 \\
entity embedding dim & 32 & 64 & 64  \\
encoder hidden dim & 48 & 32 & 32  \\
encoder num layers & 2 & 2 & 4  \\
entity hidden dim & 32 & 48 & 16  \\
system hidden dim & 32 & 32 & 32  \\
latent dim & 8 & 32 & 16  \\
regime hidden dim & 48 & 64 & 32  \\
expert hidden dim & 16 & 32 & 32  \\
$\tau_{max}$ & 1.2 & 1 & 1  \\
$\tau_{min}$ & 0.5 & 0.3 & 0.5  \\
\midrule
prediction weight & 5.0 & 5.0 & 2.0  \\
prior prediction weight & 0.5 & 0.5 & 0.5  \\
dynamics kl weight & 0.8 & 0.8 & 0.8  \\
representation kl weight & 0.2 & 0.8 & 0.2  \\
structural weight & 2.0 & 1.0 & 2.0  \\
entropy weight & 0.2 & 0.05 & 0.05  \\
orthogonality weight & 0.05 & 0.01 & 0.01  \\
\bottomrule
\end{tabular}
\caption{\textbf{Hyperparameters for all datasets.} List of the relevant hyperparameters and their values per dataset.}
\label{tab:hyper}
\end{table}

\subsection{Model Size and Computational Budget}
\label{app:params}

Table~\ref{tab:params} reports trainable parameter counts for the implemented models and configurations. Parameter counts are provided to make clear that the forecasting results are not obtained by comparing HERON only against much smaller baselines.

\begin{table}[ht]
\captionsetup{skip=10pt}
\centering
\small
\begin{tabular}{lll}
\toprule
Model & Dataset/configuration & Trainable parameters \\
\midrule
GraphSSM & Synthetic & 300K \\
GraphSSM & NBA & 450K \\
GraphSSM & AMASS & 450K \\
\midrule
HERON & Synthetic & 400K \\
HERON & NBA & 600K \\
HERON & AMASS & 600K \\
\midrule
EvolveGCN & Synthetic & 300K \\
EvolveGCN & NBA & 800K \\
EvolveGCN & AMASS  & 800K \\
\midrule
S4 & Synthetic & 600K \\
S4 & NBA  & 1.7M \\
S4 & AMASS  & 1.7M \\
\midrule
LaM-SLidE & Synthetic  & 1.7M \\
LaM-SLidE & NBA & 2.1M \\
LaM-SLidE & AMASS & 2.1M \\
\midrule
Persistence & Any dataset & 0 \\
\bottomrule
\end{tabular}
\caption{\textbf{Trainable parameter counts.} Counts correspond to the instantiated configurations used in the experiments.}
\label{tab:params}
\end{table}

Training was performed on a single A100 40GB GPU using four data-loading workers. Depending on the dataset, training took approximately 10 to 90 minutes.

\paragraph{Cost of expert specialization.}
Each expert is a small head (3-layer MLPs, the full $K$-expert bank is around 12\% of the model). All $K$ experts are evaluated at every step regardless of the decorrelation term, because the transition forms a straight-through mixture $\sum_k w_k\,\mathrm{expert}_k$ over experts; their outputs are already materialized in the forward pass. The decorrelation regularizer therefore adds no extra expert evaluations; it only reads those existing outputs and computes a $K\times K$ similarity, which is negligible. The cost of evaluating all experts rather than a single hard-selected one is a lightweight-head overhead of approximately 7\% wall-clock on GPU.

\subsection{Limitations}
\label{app:limitations}

\subsubsection{\texorpdfstring{MoE scales linearly with $K$.}{MoE scales linearly with K.}}

Despite its alignment with our problem formulation as explained in section \ref{app:moe}, a limitation of the choice of MoE as the modulation mechanism is that the number of expert parameters scales linearly with the number of regimes $K$. This scaling is not specific to our framework, but follows from explicitly representing $K$ separate transition mechanisms. When $K$ is moderate and the regimes have interpretable semantics, this cost is acceptable and provides the benefit of separately parameterized dynamics. However, for settings with a large number of regimes, or with regimes that vary more continuously, lighter conditioning mechanisms may be preferable. One possible alternative is FiLM-style modulation~\citep{film}, where a shared transition model is modulated by the regime variable. Such an approach would preserve regime-dependent dynamics while largely decoupling the size of the main transition model from $K$. The specialization machinery used in our framework could in principle be retained, since the corresponding objectives are defined on expert or transition outputs rather than on a specific internal parameterization.

This alternative comes with a trade-off. FiLM-style modulation does not provide separately parameterized experts, and therefore cannot directly encode different architectural inductive biases across regimes. For example, one transition function could be parameterized as a GNN while another could use a Transformer architecture~\citep{transformer}. Similarly, when the form of the dynamics is partially known, or when a known basis of transition functions is available for modeling flows on the network, an explicit MoE parameterization remains a natural choice. We therefore view the MoE design as appropriate for the regime-structured setting considered in this work, while leaving a systematic comparison with lighter modulation mechanisms to future work.

\subsubsection{Losses requires balance between weights.}
Beyond the prediction objective, our training objective includes terms that encourage expert usage, expert specialization, prior--posterior alignment, and prevention of posterior collapse. These terms introduce additional hyperparameters, and the behavior of the model depends on their relative weighting. If the usage or specialization terms are too weak, the model may fail to assign distinct dynamics to different experts. Conversely, if these terms dominate the predictive objective, specialization can be encouraged at the expense of accurate forecasting. Similarly, insufficient prior--posterior alignment can lead to unreliable open-loop rollouts, while excessive pressure on the posterior may contribute to collapse. These failure modes are well known in latent dynamics models and structured representation learning~\citep{dreamerv2,dreamerv3,structureloss}.

We include several safeguards to reduce this sensitivity. First, we adopt the KL balancing mechanism introduced in~\citep{dreamerv2}, which stabilizes prior--posterior alignment by discouraging the posterior from moving into regions that the prior cannot reach. Second, in our experiments we observed that the structural objectives require primarily a stable relative balance rather than independent per-dataset tuning. In particular, a fixed ratio between the expert-usage and specialization terms was effective across the datasets considered, with only global scaling needed to balance these structural losses against the prediction loss. From Section \ref{app:hyperparameters}, a ratio of $\frac{1}{4}$ is constant over the entropy and orthogonality weights, where the structural weights tune generally the constraint put on the model to restrict its latent space. This is consistent with the role of these terms: they jointly encourage peaked assignments at the sample level while maintaining balanced expert usage at the population level which naturally comes with a tradeoff and balance. We also observed limited sensitivity around the $0.8/0.2$ KL-balancing weights used in our experiments. They were consistently selected as best threshold in the hyperparameters search \ref{app:hyperparameters}.

These observations do not remove the need for validation when transferring the framework to substantially different domains, larger values of $K$, or noisier regimes. The loss weights remain hyperparameters, and incorrect balancing can still lead to underused experts, weak specialization, or inaccurate rollouts. In practice, these issues can be monitored through expert-usage statistics, assignment entropy, specialization metrics, and prior--posterior KL values. A more exhaustive study of automatic loss-weight adaptation is left for future work.

\subsubsection{Uniform regime usage is restrictive under severe imbalance.}

The submitted uniform balance objective is unsuitable under severe imbalance: minority recall falls to zero in the controlled setting of Appendix~\ref{app:imbalance}. Replacing its fixed target with an unsupervised learned categorical prior recovers minority recall to 0.70 and improves macro-F1 from 0.60 to 0.78. We therefore view the learned-prior formulation as an extension and explicitly retain the limitation of the original uniform objective.

\subsubsection{One active regime at rollout.}

The current hard rollout gate assumes that one regime is active at each time step. The controlled experiment in Appendix~\ref{app:multik} shows that HERON's posterior already captures mixed regime membership, while the limitation arises from collapsing it to one expert at rollout. Posterior-weighted routing provides a straightforward extension for settings with coexisting dynamics. This experiment addresses simultaneous mixtures; continuous temporal transitions remain approximated under the current discrete prior.

\subsubsection{Fixed target topology.}

HERON currently models regime-dependent dynamics on a fixed target topology and does not generate node/edge birth or death events. Topology-changing generation is outside the scope of this work, but can be envisioned as an interesting future extension

\subsubsection{\texorpdfstring{Number of regimes $K$ is a model-selection hyperparameter.}{Number of regimes K is a model-selection hyperparameter.}}

HERON treats the number of regimes $K$ as a fixed model-selection hyperparameter. When regime labels or domain knowledge are unavailable, Appendix~\ref{app:k} selects $K$ using the predictive NLL of held-out validation futures and a fixed elbow rule. This recovers the reference $K=3/2/4$ on Network/NBA/AMASS, whereas the velocity heuristic selects 2/4/5. The sensitivity analysis in Appendix~\ref{app:k_sensitivity} further shows that forecasting is relatively robust to moderate over-specification, whereas reference-label-aligned recovery remains sensitive to both under- and over-specification. Fully adaptive regime discovery, where new regimes can emerge or be pruned during training, remains a natural direction for future work.

\clearpage
\section*{NeurIPS Paper Checklist}

\begin{enumerate}

\item {\bf Claims}
    \item[] Question: Do the main claims made in the abstract and introduction accurately reflect the paper's contributions and scope?
    \item[] Answer: \answerYes{} 
    \item[] Justification: All the claims found in the abstract and introduction are supported by the results  discussed in the paper.

\item {\bf Limitations}
    \item[] Question: Does the paper discuss the limitations of the work performed by the authors?
    \item[] Answer: \answerYes{} 
    \item[] Justification: We discuss all the limitations and future work in the dedicated section \ref{sec:conclusion} and Appendix ~\ref{app:limitations}.

\item {\bf Theory assumptions and proofs}
    \item[] Question: For each theoretical result, does the paper provide the full set of assumptions and a complete (and correct) proof?
    \item[] Answer: \answerYes{} 
    \item[] Justification: The paper provides a set of detailed definitions that motivate the choice of the developed model. We provide all the necessary information in the Appendix.

    \item {\bf Experimental result reproducibility}
    \item[] Question: Does the paper fully disclose all the information needed to reproduce the main experimental results of the paper to the extent that it affects the main claims and/or conclusions of the paper (regardless of whether the code and data are provided or not)?
    \item[] Answer: \answerYes{} 
    \item[] Justification: The datasets we use are publicly available and we provide details on how to generate the synthetic data in the Appendix. To make our experiments reproducible, the set of all hyperparameters of the models used along with the training specifications are provided in detail in the Appendix.

\item {\bf Open access to data and code}
    \item[] Question: Does the paper provide open access to the data and code, with sufficient instructions to faithfully reproduce the main experimental results, as described in supplemental material?
    \item[] Answer: \answerYes{} 
    \item[] Justification: Yes, the full code with detailed instructions for reproducibility have been made public and can be found at  \url{https://github.com/JerBaf/HERON}. 

\item {\bf Experimental setting/details}
    \item[] Question: Does the paper specify all the training and test details (e.g., data splits, hyperparameters, how they were chosen, type of optimizer) necessary to understand the results?
    \item[] Answer: \answerYes{} 
    \item[] Justification: We provide detailed descriptions of all the training parameters, procedure and hyperparameters used for every model that we train in the Appendix.

\item {\bf Experiment statistical significance}
    \item[] Question: Does the paper report error bars suitably and correctly defined or other appropriate information about the statistical significance of the experiments?
    \item[] Answer: \answerYes{} 
    \item[] Justification: We provide all our results in terms of downstream task performance in the format "mean $\pm$ standard deviation from the mean".

\item {\bf Experiments compute resources}
    \item[] Question: For each experiment, does the paper provide sufficient information on the computer resources (type of compute workers, memory, time of execution) needed to reproduce the experiments?
    \item[] Answer: \answerYes{} 
    \item[] Justification: We provide detailed information about the used resources in the Appendix.
    
\item {\bf Code of ethics}
    \item[] Question: Does the research conducted in the paper conform, in every respect, with the NeurIPS Code of Ethics \url{https://neurips.cc/public/EthicsGuidelines}?
    \item[] Answer: \answerYes{} 
    \item[] Justification: We preserve anonymity in our submission and our submission abides by the NeurIPS Code of Ethics.

\item {\bf Broader impacts}
    \item[] Question: Does the paper discuss both potential positive societal impacts and negative societal impacts of the work performed?
    \item[] Answer: \answerYes{} 
    \item[] Justification: We discuss potential positive and negative societal impacts in Section \ref{broader_impacts}.
    
\item {\bf Safeguards}
    \item[] Question: Does the paper describe safeguards that have been put in place for responsible release of data or models that have a high risk for misuse (e.g., pre-trained language models, image generators, or scraped datasets)?
    \item[] Answer: \answerNA{} 
    \item[] Justification:  The release of our data and models upon acceptance, does not pose a direct risk of misuse.

\item {\bf Licenses for existing assets}
    \item[] Question: Are the creators or original owners of assets (e.g., code, data, models), used in the paper, properly credited and are the license and terms of use explicitly mentioned and properly respected?
    \item[] Answer: \answerYes{} 
    \item[] Justification: We provide citations for all sources of code and/or data used. For the datasets, we reference the original open-source work they are based on, in full accordance with their licensing terms.

\item {\bf New assets}
    \item[] Question: Are new assets introduced in the paper well documented and is the documentation provided alongside the assets?
    \item[] Answer: \answerNA{} 
    \item[] Justification:  This paper does not release any new assets beyond code, which is adequately documented in order to reproduce all results from the paper.

\item {\bf Crowdsourcing and research with human subjects}
    \item[] Question: For crowdsourcing experiments and research with human subjects, does the paper include the full text of instructions given to participants and screenshots, if applicable, as well as details about compensation (if any)? 
    \item[] Answer: \answerNA{} 
    \item[] Justification: No crowdsourcing or human subjects were involved in the experiments conducted for this paper.

\item {\bf Institutional review board (IRB) approvals or equivalent for research with human subjects}
    \item[] Question: Does the paper describe potential risks incurred by study participants, whether such risks were disclosed to the subjects, and whether Institutional Review Board (IRB) approvals (or an equivalent approval/review based on the requirements of your country or institution) were obtained?
    \item[] Answer: \answerNA{} 
    \item[] Justification: No crowdsourcing or human subjects were involved in the experiments conducted for this paper.

\item {\bf Declaration of LLM usage}
    \item[] Question: Does the paper describe the usage of LLMs if it is an important, original, or non-standard component of the core methods in this research? Note that if the LLM is used only for writing, editing, or formatting purposes and does \emph{not} impact the core methodology, scientific rigor, or originality of the research, declaration is not required.
    \item[] Answer: \answerNA{} 
    \item[] Justification: LLMs are not a core component of this paper in terms of methods.

\end{enumerate}
\end{document}